\documentclass[a4paper,fleqn]{cas-dc}

\usepackage[numbers]{natbib}
\usepackage[dvipsnames]{xcolor}
\usepackage{bm}
\usepackage{graphicx}
\usepackage{grffile}
\usepackage{algorithmic}
\usepackage{algorithm}
\def\tsc#1{\csdef{#1}{\textsc{\lowercase{#1}}\xspace}}
\tsc{WGM}
\tsc{QE}
\tsc{EP}
\tsc{PMS}
\tsc{BEC}
\tsc{DE}

\begin{document}
\let\WriteBookmarks\relax
\def\floatpagepagefraction{1}
\def\textpagefraction{.001}

\shorttitle{Unified Dimensional Reduction Neural Network}

\shortauthors{Zelin Zang et~al.}

\title [mode = title]{
  UDRN: Unified Dimensional Reduction Neural Network for Feature Selection and Feature Projection
  }
\tnotemark[1]

\tnotetext[1]{Zelin Zang and Yongjie Xu contribute equally.}

% \tnotetext[2]{This document is the results of the research
%   project funded by the National Science Foundation.}

% \tnotetext[3]{The second title footnote which is a longer text matter
%   to fill through the whole text width and overflow into
%   another line in the footnotes area of the first page.}

\author[1,2,3]{Zelin Zang}[type=editor,
  style=chinese,
  bioid=1,
  auid=000,
  prefix= ,
  % role=Researcher,
  orcid=0000-0003-2831-5437]
\fnmark[1]
\author[1,2,3]{Yongjie Xu}[type=editor,
  style=chinese,
  bioid=1,
  auid=000,
  %  prefix=Sir,
  % role=Researcher,
  % orcid=0000-0001-7511-2910
]
\author[4]{Linyan Lu}[type=editor,
  style=chinese,
  bioid=1,
  auid=000,
  %  prefix=Sir,
  % role=Researcher,
  % orcid=0000-0001-7511-2910
]
\author[2]{Yulan Geng}[type=editor,
  style=chinese,
  bioid=1,
  auid=000,
  %  prefix=Sir,
  % role=Researcher,
  % orcid=0000-0001-7511-2910
]
\author[2]{Senqiao Yang}[type=editor,
  style=chinese,
  bioid=1,
  auid=000,
  %  prefix=Sir,
  % role=Researcher,
  % orcid=0000-0001-7511-2910
]
\author[2,3]{Stan.Z Li}[type=editor,
  style=chinese,
  bioid=1,
  auid=000,
  prefix=Prof,
  % role=Researcher,
  % orcid=0000-0001-7511-2910
]
\cormark[1]

\credit{Conceptualization of this study, Methodology}
\affiliation[1]{
  organization={Zhejiang University},
  % addressline={Yuquan Campus, Zhejiang University}, 
  city={Hangzhou},
  postcode={310000},
  country={China}
}
\affiliation[2]{
  organization={Westlake University},
  addressline={AI Lab, School of Engineering},
  city={Hangzhou},
  postcode={310000},
  country={China}
}
\affiliation[3]{
  organization={Westlake Institute for Advanced Study},
  addressline={Institute of Advanced Technology},
  city={Hangzhou},
  postcode={310000},
  country={China}
}
\affiliation[4]{
  organization={China Telecom Corporation Limited, Hangzhou Branch},
  % addressline={Institute of Advanced Technology},
  city={Hangzhou},
  postcode={310000},
  country={China}
}

\shortauthors{Zelin Zang et~al.}

\begin{abstract}
  Dimensional reduction~(DR) maps high-dimensional data into a lower dimensions latent space with minimized defined optimization objectives. The two independent branches of DR are feature selection~(FS) and feature projection~(FP). FS focuses on selecting a critical subset of dimensions but risks destroying the data distribution~(structure). On the other hand, FP combines all the input features into lower dimensions space, aiming to maintain the data structure; but lacks interpretability and sparsity. 
  Moreover, FS and FP are traditionally incompatible categories and have not been unified into an amicable framework. Therefore, we consider that the ideal DR approach combines both FS and FP into a unified end-to-end manifold learning framework, simultaneously performing fundamental feature discovery while maintaining the intrinsic relationships between data samples in the latent space. 
  This paper proposes a unified framework named Unified Dimensional Reduction Network~(UDRN) to integrate FS and FP in an end-to-end way. Furthermore, a novel network framework is designed to implement FS and FP tasks separately using a stacked feature selection network and feature projection network. In addition, a stronger manifold assumption and a novel loss function are proposed. Furthermore, the loss function can leverage the priors of data augmentation to enhance the generalization ability of the proposed UDRN\@. Finally, comprehensive experimental results on four image and four biological datasets, including very high-dimensional data, demonstrate the advantages of DRN over existing methods~(FS, FP, and FS\&FP pipeline), especially in downstream tasks such as classification and visualization.
\end{abstract}

% \includegraphics{figs/grabs.pdf}

% \begin{highlights}
%   \item \textbf{Unified FS\&FP problem and a novel neural network framework:} We propose the problem of FS\&FP with a unified objective and design a neural network framework to solve this problem. The problem is widely needed in biology, genomics, and proteomics.
%   \item \textbf{Novel data augmentation for FS\&FP and corresponding loss functions:} We propose a novel structure-preserving loss function and online FMH augmentation to provide a consistent and generalizable objective function.
%   \item \textbf{Extensive experiments and promising results:} We compare DRN with state-of-the-art FS, FP, and FS\&FP pipeline methods on ten datasets and a case study of the supervised scenarios.
% \end{highlights}

% Keywords

\maketitle

\section{Introduction}
Dimensional reduction~(DR)~\cite{ayesha2020overview, li2020deep, ZangEVNet2022} transforms a high-dimensional~(h-dim) data into an intrinsic low-dimensional~(l-dim) embedding. 
The performance of typical classification or visualization methods degrades when data has too many features. Therefore, DR is introduced to overcome this issue. DR have a broad range of applications in signal processing~\cite{chen2013efficient}, speech recognition~\cite{liu2018speech}, neuroinformatics~\cite{deraeve2018fast}, and bioinformatics~\cite{remeseiro2019review}.

The ideal DR method is expected to have two \emph{characteristics}~\cite{van2009dimensionality}.
(1) \textbf{Structural maintainability.} The local structures of the data need to be preserved from being broken while reducing the data dimensionally. Under the manifold assumption, ensuring the local connectivity of the data is the golden rule for structure preservation.
(2) \textbf{Sparse interpretability.} Redundant features and noisy features need to be identified while reducing the data dimensionally. It is because these useless features can affect the accuracy of downstream tasks.

In many biology exploration fields, such as single-cell~analysis~\cite{wang2021single, sheng2021selecting, zhang2021critical} genomics~\cite{jiang2022statistics} and proteomics~\cite{ sun2022artificial}, DR is required to have both characteristics. 
However, the current DR methods cannot achieve the above \emph{characteristics} with a unified framework. The current DR methods often contain two incompatible branches, feature projection~(FP)~\cite{Xia2021Revisiting} and feature selection~(FS)~\cite{wei2016unsupervised}.
FP methods concentrate on \textbf{structural maintainability}. They produce new variables obtained from the original features via an arbitrarily complex mapping, thus having better distinguish and structure-preserving performance.
In contrast, FS methods concentrate on \textbf{sparse interpretability}. They allow the user to find an essential feature~subset during the generalization phase of the model but often break the structure of the data by losing features~\cite{li2020ivfs, Abubakar, XinxingWu2021FractalAF}.

Due to the needs of practical fields such as biology exploration, some researchers build pipeline methods by splicing the FS and FP methods together to solve the above issues~\cite{townes2019feature, zhang2021critical, sun2022artificial}.
{
\color{black} However, these pipeline methods are not satisfying due to the following reasons.
\textbf{(1) Inconsistent optimization objective.} The stacked approach of FS and FP may introduce conflicting objectives, e.g., FS focuses on reconstruction error while FP focuses on distance/similarity preserving. The DR method that can combine both FS and FP has not been found yet.
\textbf{(2) Weak generalizability.} FS and FP are mainly applied in scenarios with a huge number of features, where it is easy to fall into overfitting due to the relatively small number of samples.
% \textbf{(3) Feature information leakage.} Almost all NN-based FS approaches take offline feature selection, implying that score optimization and subset selection are performed separately, resulting in FS not being embedded in the end-to-end network.
% When scoring optimization, much information about unimportant features that should be discarded is leaked into the network, affecting the performance of feature selection.
}

{
\color{black}
In this paper, an end-to-end Unified Dimensional Reduction Network, named UDRN, is designed to perform feature selection and projection~(FS\&FP) in a unified framework.
\textbf{To design consistent optimization objective}, a novel neural network framework is designed. The framework includes a feature~selection~network~(FS network) and feature~projection~network~(FP network) for both FS and FP tasks. The gate layer of the FS network \emph{mask off} unimportant features and generate feature~subset during the forward propagation process, thus implementing online feature selection and enabling the following FP with selected features. The FP network then maps the feature~subset to l-dim space for downstream tasks such as classification and visualization.
\textbf{To improve the generalizability of UDRN}, a manifold~connectivity~\cite{lin2008riemannian,mcinnes_umap_2018} based manifold assumption using priors of data augmentation is proposed. Furthermore, based on the above assumption, a novel loss function is designed to train the FS\&FP neural network with online generated augmented data. The proposed unsupervised loss function is compatible with the new data generated by data augmentation, thus allowing a finer depiction of the data manifold and ultimately leading to improved performance.
}

{
    \color{black}
    To the best of our knowledge, UDRN is the first attempt to apply data augmentation-compatible structure-preserving loss functions on the neural network for the FS\&FP task. Our contributions are summarized as follows.
}
\begin{itemize}
    \item {\color{black} \textbf{A unified FS\&FP task and a novel neural network framework.} {We explicitly define the FS\&FP task and design a neural network framework with a unified objective to solve this task.} The proposed task is extensively employed in biology, genomics, and proteomics.}
    \item \textbf{A manifold connectivity assumption under augmented data and corresponding novel loss function.} We propose a novel structure-preserving loss function and online data augmentation to provide a consistent and generalizable objective function.
    \item \textbf{Extensive experiments and promising results.} We compare UDRN with state-of-the-art FS, FP, and FS\&FP pipeline methods on ten datasets and a case study of the supervised scenarios.
\end{itemize}
\section{Related Work}
\label{sec_relatedwork}

\subsection{Feature Selection~(FS)}

The FS methods include four categories~\cite{alelyani2018feature},
(a) filter methods, which are independent of learning models;
(b) wrapper methods, which rely on learning models for selection criteria;
(c) embedder methods, which embed the FS into learning models to also achieve model fitting simultaneously;
(d) hybrid approaches, which are a combination of more than one of the above three.
% Alternatively, these methods are categorized as supervised and semi-supervised methods.
{
\color{black}
Unsupervised FS is more widely used because it does not require information about the label. At the same time, unsupervised FS is more challenging due to the same reasons~\cite{peng2017nonnegative, NattaneLuizadaCosta2021EvaluationOF}.
}

From another perspective, FS methods can be divided into non-parametric and parametric methods.
\textit{Non-parametric models select the appropriate features based on statistics.} For example, Laplacian score~(LS)~\cite{he2006laplacian} uses the nearest neighbor graph to model the local geometric structures of the data. 
Principal feature analysis~(PFA)~\cite{lu2007feature} utilizes the structure of the main components of a set of features to select the subset of relevant features. 
Multi-cluster feature selection~(MCFS)~\cite{cai2010unsupervised} selects a subset of features to cover the multi-cluster structure of the data, where spectral analysis is used to find the inter-relationship between different components.
In Unsupervised discriminative feature selection~(UDFS)~\cite{yang2011l2}, the discriminatory analysis method and $\ell_{2,1}$~regularization are used to identify the valuable features.
Nonnegative~discriminative~feature~selection~(NDFS)~\cite{li2012unsupervised} select discriminative features by learning the cluster labels and FS matrix.
The NDFS uses a nonnegative constraint on the class indicator to understand cluster labels and adopts an $\ell_{2,1}$ limitation on the redundant features.
IVFS~\cite{li2020ivfs} select useful features by preserving the pairwise distances, as well as topological patterns, of the complete data.

\textit{Parametric models select the appropriate features based on neural networks.}
For example, Autoencoder Feature Selector (AEFS)~\cite{Han} combines reconstruction loss and $\ell_{2,1}$ regularization loss to obtain a subset of useful features on the weights of the encoder. 
The agnostic feature selection (AgnoS)~\cite{Doquet} combines AE with different auxiliary tasks to design a range of FS methods.
Such as AgnoS-W\@: the $\ell_{2,1}$ norm on the weights of the first layer of AE,
AgnoS-G\@: $\ell_{2,1}$ norm on the gradient of the encoder, and AgnoS-S\@: $\ell_1$ norm on the slack variables that constitute the first layer of AE\@.
Concrete Autoencoders~(CAE)~\cite{Abubakar} replaces the first hidden layer of AE with a concrete selector layer, which is the relaxation of a discrete distribution called concrete distribution~\cite{Maddison}.
Fractal Autoencoders~(FAE)~\cite{XinxingWu2021FractalAF} extends autoencoders by adding a one-to-one scoring layer. FAE uses a small sub-neural network for FS in an unsupervised fashion.
{\color{black}
Atashgahi et.al~\cite{atashgahi2021quick} introduce the strength of the neuron in sparse neural networks as a criterion to measure the feature importance and designs QuickSelection~(QS).

We consider that it is a meaningful research direction for designing FS methods based on neural networks. Training FS models based on reconstruction loss cannot take into account structure preservation \& feature projection; thus, it is meaningful to design novel structure-preserving loss functions for both FS and FP tasks.
}

\subsection{Feature Projection~(FP)}

In recent years, Numerous manifold-learning-based FP methods have been proposed.
Some of the FP methods are based on the manifold assumption~\cite{Belkin-Niyogi-03, Fefferman-manifold-2016}, which states that a pattern of interest in data is a lower-dimensional manifold~(or hyper-surface) residing in the high dimensional data space.
When the data contains multiple manifolds, the geometric structure usually includes the local system of neighboring points on each manifold and global relationships among different manifolds.

FP methods can be divided into non-parametric and parametric methods.
In terms of non-parametric methods, Isometric Mapping (ISOMAP)~\cite{Tenenbaum-science-00} and Locally Linear Embedding (LLE)~\cite{Roweis-science-00} are classic ones, among others.
Later developments include Hessian LLE (HLLE)~\cite{donoho2003hessian}, Modified LLE~(MLLE)~\cite{zhang2007mlle}.
The t-Distributed Stochastic Neighbor Embedding~(t-SNE)~\cite{Maaten-t-SNE-JMLR-2008} and Uniform Manifold Approximation and Projection~(UMAP)~\cite{mcinnes2018umap} are two popular methods for manifold learning-based Nonlinear dimensionality reduction~(NLDR), widely used for NLDR and visualization.
The t-SNE improves the previous work of SNE~\cite{SNE-2007} by using a long-tailed $t$-distribution for the embedding layer~\cite{Maaten-t-SNE-JMLR-2008}.
The UMAP further introduces a global term added to the local neighborhood-based t-SNE to preserve the global structure.

{
    \color{black}
    In terms of parametric methods, the Deep Isometric Manifold Learning~(DIMAL)~\cite{pai2018dimal} combines a deep learning framework with a multi-dimensional scaling~(MDS) objective, which can be seen as a neural network version of MDS\@. DIMAL learns distance-preserving mapping to generate low-dimensional embeddings for a particular class of manifolds with sparse geodesic sampling. Topological Autoencoder~(TAE)~\cite{moor2021topological} imposes topological constraints~\cite{Wasserman-2018} on top of the autoencoder architecture to preserve the topological structure of data. Sainburg et.al~\cite{TimSainburg2021ParametricUL} extend the embedding step of UMAP~\cite{mcinnes2018umap} to a parametric optimization over neural network weights, learning a parametric relationship between data and embedding. 
    DLME~\cite{zang2022dlme} is a generalizable neural network with manifold flatness assumption which can handle biological and image data well.
}

\subsection{FS and FP~(FS\&FP) Piplines}

{
    \color{black}
    Some pipeline methods are designed to combine feature selection and feature projection in fields such as bioinformatics~(in single-cell~analysis~\cite{MalteDLuecken2019CurrentBP, wang2021single, sheng2021selecting, zhang2021critical}, genomics~\cite{AndriesTMarees2018ATO, jiang2022statistics}, and proteomics~\cite{ChristinaLudwig2018DataindependentAS, sun2022artificial}). 
    The pipeline method includes an FS method, which discovers the significant features, and an FP method, which analyzes the effects of these features on phenotype. Since no corresponding end-to-end FS and FP analysis methods can be found, such tasks are often performed using a pipeline approach. For example, different FS and FP methods are used in series to complete the analysis task~\cite{sun2021single, mann2021artificial, kustatscher2022understudied}.

    As demonstrated above, the pipeline approach may cause corruption of helpful information due to the non-uniform loss functions and interrupted information flow of FS and FP\@. Therefore, we innovatively propose UDRN based on a neural network, which accomplishes feature selection and projection through an end-to-end network framework.
}

\section{Problem Defenition and Preliminaries}
\subsection{Problem Definition}
We use bold uppercase characters for matrices (e.g., $\bm{A}$), bold lowercase characters for vectors (e.g., $\bm{a}$), and regular lowercase characters for scalars (e.g., $a$).
Also, we represent the $i$-th element of vector $\bm{a}$ as $a_i$, the $i$-th row of matrix $\bm{A}$ as $A_{i*}$, the $j$-th column of matrix $\bm{A}$ as $A_{*j}$, the $(i, j)$-th entry of matrix $\bm{A}$ as $A_{ij}$, the transpose of $\bm{A}$ as $\bm{A}^T$.
{\color{black} We introduce our proposed concept on the attributed graph~\cite{pfeiffer2014attributed} to precisely describe our data augmentation and loss function and adapt it to a broader range of situations.}

\noindent \textbf{Definition 3.1 (Attributed Graph)}. Let graph $\bm{\mathcal{G}}(\bm{\mathcal{V}}, \bm{\mathcal{E}}, \bm{\mathcal{X}})$ be an attributed graph~(network). {\color{black} It consists of -
        (1) $\bm{\mathcal{V}}$, the set of nodes, $n=|\bm{\mathcal{V}}|$, where $n$ is the number of the nodes.
        (2) $\bm{\mathcal{E}}$, the set of edges, $e=|\bm{\mathcal{E}}|$, where $e$ is the number of the edges.}
and (3) $\bm{\mathcal{X}}=[\bm{x}_1, \bm{x}_2, \ldots, \bm{x}_n]$, the set of node attributes (features),
where $\bm{x}_i\in \mathcal{R}^\mathcal{D}$, $\mathcal{D}$ is the dimensional number of attribute.

With the definition of the attributed graph, we now define the FS\&FP problem as follows.

\noindent \textbf{Definition 3.2 (FS\&FP Task on Attributed Graph)}.
{
\color{black}
Given an attributed network $\bm{\mathcal{G}}(\bm{\mathcal{V}}, \bm{\mathcal{E}}, \bm{\mathcal{X}})$, the FS\&FP task on attributed graph aims to
(a) select a subset of $d$ features from the original $\mathcal{D}$-dimensional feature space, and $d\ll \mathcal{D}$ and then
(b) \color{black} map the data with selected features $\bm{\mathcal{X}}^s$ to a latent space $\bm{\mathcal{Z}}$.
We expect that the selected feature subsets and the generated embedding representations imply as much information as possible as the original data. It may manifest itself in as high an accuracy as possible in downstream tasks, neighborhood structure maintenance, and consistent visualization output.

The adjacency of nodes is described by the edges of the attribute graph, which contains critical a priori knowledge. In the unsupervised context, $k$-NN is used to build the edge structure.}
\begin{equation}
    \bm{\mathcal{E}}=\{
    (\bm{v}_i, \bm{v}_j) | \bm{v}_j \in \mathcal{N}^k(\bm{v}_i), \bm{v}_i \in \bm{\mathcal{V}}
    \}
    \label{eq_unsup}
\end{equation}
{
\noindent {\color{black} where $\mathcal{N}^k(\bm{v}_i)$ }is set of $k$-NN neighborhood of node $\bm{v}_i$, $k$ is the hyperparameter of $k$-NN\@.
\color{black}
}
UDRN can be easily compatible with the supervised situation because the methods and problems are based on attribute graphs. When additional supervised information or structural information is accessible, we only need to redefine the edges.
\begin{equation}
    \bm{\mathcal{E}}=\{
    (\bm{v}_i, \bm{v}_j) | \bm{v}_j \in \mathcal{N}^k(\bm{v}_i) \cap Y(\bm{v}_i) , \bm{v}_i \in \bm{\mathcal{V}}
    \}
    \label{eq_sup}
\end{equation}
where $Y(\bm{v}_i)$ is the set of nodes with the same label as $\bm{v}_i$.

\subsection{AE-based Feature Selection}

AE-based FS methods add a superficial one-to-one layer between the input and hidden layers, which can weigh the importance of each feature~\cite{Han, Abubakar, XinxingWu2021FractalAF}.
{\color{black} The parameters of the one-to-one layer are trained by $L_1$ regularization loss and reconstruction loss, thus these methods highlight the features which friendly to reconstruction.} The loss function of AE-based FS is,
\begin{equation}
    \begin{aligned}
        L_\text{fp} = \min _{
            \bm{\mathcal{W}}, f, g
        }
        \left\|{ \bm{x} }-g
        \left(
        f\left(\bm{x} \bm{\mathcal{W}}\right)
        \right)
        \right\|_{\mathrm{F}}^{2}
        +
        \lambda_{1}
        {L}_1(\bm{w})
    \end{aligned}
    \label{eq:traditionFS}
\end{equation}
{\color{black}
where $\bm{\mathcal{W}}$ is a trainable parameter matrix with values only on the diagonal. $\bm{\mathcal{W}}=\text{Diag}(\bm{w}) \in \mathcal{R}^{\mathcal{D}\times \mathcal{D}}$, and the $w_j$ is the importance of feature $j$,
$\bm{w} = \left\{w_j | j \in \left\{1, 2, \ldots, \mathcal{D}\right\} \right\} \in \mathcal{R}^{\mathcal{D}}$~(check Eq.~(3) in \cite{wu_fractal_2020} for more details).
The $\left\|\cdot\right\|_{\mathrm{F}}$ is the Frobenius norm~\cite{bottcher2008frobenius}.
% The $g_\phi$ and $f_\theta$ are the  and decoder. 

AE-based FS methods include encoder $f(\cdot)$ and decoder $g(\cdot)$.}
The encoder $f(\cdot)$ embeds the input data $\bm{x} \bm{\mathcal{W}}$ into a latent space, and the decoder $g(\cdot)$ maps the latent space data back to the original space and calculates the reconstruction loss.
The $L_1$ loss leads to a decrease in $\bm{w}$. The reconstruction loss increases the feature importance of important features, and the two losses act synergistically to guide important features to have higher scores.

Most current FS methods involve an \emph{offline feature selection} strategy. It includes two steps:
(1) all features are scored for importance using various objective functions;
(2) the top-k essential features are selected.
The above offline scheme poses an obstacle to the unified FS\&FP task, for two reasons.
(D1) \textbf{Leakage of unimportant feature information.} During training, the forward propagation of unimportant features is not interrupted. However, the unimportant features are not accessible during inference. The above bias causes a large variance between inference results and training results.
(D2) \textbf{Difficult to preserve the structure well in the latent space.}
The feature-level reconstruction is focused on~(possibly including redundant and noisy features) without considering the neighborhood structure of the data.

{\color{black}
\subsection{Data Augmentation on Attributed Network}

Data augmentation is a commonly used NN training method for image classification and signal processing~\cite{shorten2019survey}. It acts as a regularizer and helps to reduce overfitting. We find that typical image data augmentation schemes cannot directly apply to FS\@. The reason is that FS requires the meaning of individual features is not destroyed by data augmentation. To this end, the data augmentation schemes which do not change the meaning of the feature are adapted in the FS\&FP task.

\noindent \textbf{Definition 3.3 (Data Augmentation, w.r.t $\tau$)}. Given an attributed network $\bm{\mathcal{G}}(\bm{\mathcal{V}}, \bm{\mathcal{E}}, \bm{\mathcal{X}})$,
data augmentation generates a corresponding augmented graph $\bm{\mathcal{G}}'(\bm{\mathcal{V}}', \bm{\mathcal{E}}', \bm{\mathcal{X}}')$ and for detail,
\begin{equation}
    \begin{aligned}
        \bm{\mathcal{V}}' & =\{ \bm{v}_1, \ldots, \bm{v}_n, \bm{v}_1{'}, \ldots, \bm{v}_n{'}\}                 \\
        \bm{\mathcal{E}}' & =\{ \bm{\mathcal{E}} + \bm{\mathcal{\tilde{ E}}}' + \bm{\mathcal{\tilde{E}}}{''}\} \\
        \bm{\mathcal{X}}' & =\{ \bm{x}_1, \ldots, \bm{x}_n, \tau(\bm{x}_1), \ldots, \tau(\bm{x}_n)\}
    \end{aligned}
    \label{eq_aug_all}
\end{equation}

The augmented graph $\bm{\mathcal{G}}{'}$'s node $\bm{\mathcal{V}}'$ contains:
(1) original nodes $\bm{v}_1, \ldots, \bm{v}_n$,
and (2) augmented nodes $\bm{v}_1{'}, \ldots, \bm{v}_n{'}$, which corresponds to the original node one by one.
The edges $\bm{\mathcal{E}}'$ contains:
(1) original edges $\bm{\mathcal{E}}$, which edges between the original nodes,
(2) the inter-augmented edges: $\bm{\mathcal{\tilde{E}}}{'}=\{(\bm{v}_1, \bm{v}_1{'}), \ldots, (\bm{v}_n, \bm{v}_n{'})\}$,
and (3) intra-augmented edges $\bm{\mathcal{\tilde{E}}}{''}$, which between the augmented nodes $\bm{\mathcal{\tilde{E}}}{''}=(\bm{v}_i, \bm{v}_j) \to (\bm{v}_i{'}, \bm{v}_j{'})$.
Three kinds of edges are qualitatively different and should be modeled separately. However, we focus on the nearest neighbor relationship depicted by the edge structure and model the three edge structures as homogeneous graphs for modeling convenience.

The data augmentation operator generates new data by fusing the local structure information and random distribution. Several data augmentations are as follows.

\textbf{(i) Uniform-type data augmentation (w.r.t $\tau_{U}$)}
generates augmented data by linear combination. The linear combination parameter $r_u$ is sampled from the uniform distribution $U(0, p_U)$, and $p_U$ is the hyperparameter.
\begin{equation}
    \begin{aligned}
         & \tau_{U}(\bm{x})  = (1-r_u) \cdot \bm{x} + r_u \cdot \bm{\tilde{x}}, \\
         & \bm{\tilde{x}} \sim \mathcal{N}^k(\bm{x}), r_u \sim U(0,p_U)         \\
        %    s.t. &\sum_{k=0}^K r_k = 1    
    \end{aligned}
\end{equation}
where $\bm{\tilde{x}}$ is sampled from the attributes set of data $\bm{x}$'s $k$-NN neighborhood $\mathcal{N}^k(\bm{x})$.

\textbf{(ii) Bernoulli-type data augmentation (w.r.t $\tau_{B}$)}
generates the augmented data by directly replacing the original features with the features at the corresponding positions of the adjacent data. The probability of replacement $b_j$ is sampled from the Bernoulli distribution $B(p_{B})$, and $p_{B}$ is the success probability of the Bernoulli distribution.
\begin{equation}
    \begin{aligned}
        \tau_{B}(\bm{x}) & = \bm{x} \circ \bm{b} + \bm{\tilde{x}}  \circ (1-\bm{b}), \bm{\tilde{x}}  \sim \mathcal{N}^k(\bm{x}), \\
        \bm{b}           & = \{b_j | b_j \sim B(p_{B}), j\in\{ 1, \ldots, \mathcal{D}\}\}                                        \\
    \end{aligned}
\end{equation}
where $\circ$ is the Hadamard product.

\textbf{(iii) Normal-type data augmentation (w.r.t $\tau_{N})$} generates augmented data by adding some noise, noise parameters $b_j$ is sampled from the normal distribution $N(0, p_{N})$, and $p_{N}$ is the standard deviation. The distance between neighboring sample features as a scaling factor to avoid destroying a single feature.
\begin{equation}
    \begin{aligned}
        \tau_{N}(\bm{x}) & = {\bm{x}} + (\bm{x}-\bm{\tilde{x}} ) \circ \bm{b}, \bm{\tilde{x}} \sim \mathcal{N}^k(\bm{v}), \\
        \bm{b}           & = \{b_j | r_j \sim N(0, p_{N}), j\in\{1, \ldots, \mathcal{D} \}                                \\
    \end{aligned}
\end{equation}

During network training, the data augmentation operators are applied online, thus providing more randomness of the data and guaranteeing that the feature meaning does not change. We also discuss the effect of different data augmentation on Table.~\ref{tab_parameters_analysis}.
}

\subsection{Node Similarly and vanilla DR method}

{\color{black} Node Similarly~(NS)~\cite{zang2021unsupervised, pan2010detecting} is used to describe the relationship of nodes on the graph.}

\noindent \textbf{Definition 2.3 (Node Similarly, NS, w.r.t. $\mathcal{\bm{S}}$)}.
{
\color{black}
Given an attributed network $\bm{\mathcal{G}}(\bm{\mathcal{V}}, \bm{\mathcal{E}}, \bm{\mathcal{X}})$, and a kernel function $\kappa\left(\cdot,\cdot\right)$, the node similarly between two node $v_i$ and $v_j$ is,
\begin{equation}
    \begin{aligned}
        \mathcal{\bm{S}}_{ij}^{\bm{\mathcal{G}}} & = \kappa \left(v_i, v_j\right), % d_{v_i, v_j} = {\left\|\bm{x}_i-\bm{x}_j\right\|_{2}} \\
    \end{aligned}
    \label{equ_SSG}
\end{equation}
\noindent The kernel function transforms the distance relationship between nodes into the similarity relationship and thus constructs the structure-preserving loss function.
% We often use Gaussian kernel
The typical kernel functions include Gaussian kernel~\cite{kobak_umap_2019},
\begin{equation}
    \begin{aligned}
         & \kappa^\text{Ga}\left(v_i, v_j, \sigma\right)= \frac{1}{\sqrt{2 \pi} \sigma} \exp \left( \frac{-d_{v_i, v_j}^2}{2 \sigma^{2}}\right), \\
         & d_{v_i, v_j} = {\left\|\bm{x}_i-\bm{x}_j\right\|_{2}},
    \end{aligned}
\end{equation}
where $\sigma$ is a scaling factor.
Another typical kernel function is t-kernel~\cite{Maaten-t-SNE-JMLR-2008},
\begin{equation}
    \kappa^\text{t}\left(v_i, v_j, \nu\right)=
    \frac{1}{\sqrt{\nu} \cdot \mathrm{B}\left(0.5, 0.5\nu\right)} {\left(1+\frac{d_{v_i, v_j}^2}{\nu}\right)}^{-0.5({\nu+1})}
\end{equation}
where $\nu$ is the freedom degree of t-distribution and where $B(\cdot)$ is the Beta function.

The vanilla DR~(FP) loss function first normalizes the pairwise distance of the input data and then optimizes the latent space based on the normalized pairwise distance. Here we discuss t-SNE~\cite{Maaten-tSNE-2014} as a typical example. The t-SNE, as well as UMAP~\cite{kobak_umap_2019}, calculates the scaling factor $\sigma_i^*$ for every single node $i$ by binary search.
\begin{equation}
    \begin{aligned}
        \sigma_i^*   & = \arg\min_{\sigma_i} \left\| -\sum_{j}  p_{j \mid i} \log_2 p_{j \mid i}  -\log _{2}(p)\right\|,                                                \\
        % \left\| \sum_{j} \frac{1}{\sqrt{2 \pi} \sigma_i^2} \exp \left( \frac{-d_{v_i, v_j}}{2\sigma_{i}^2}\right)-\log _{2}(p)\right\|
        p_{j \mid i} & =\frac{\kappa^\text{Ga}\left(d_{v_i, v_j}^\text{input}, \sigma\right)}{\sum_{k \neq i} \kappa^\text{Ga}\left(d_{v_i, v_k}^\text{input}, \sigma\right)}
    \end{aligned}
\end{equation}
where the hyper-parameter `perplexity' $p$ controls the above cost function, $d_{v_i, v_j}^\text{input}$ is the distance if $i$ and $j$ in the input space, $p_{j \mid i}$ is the conditional probability. Next, t-SNE minimizes the difference between the input and latent space using Kullback Leibler divergence~\cite{SolomonKullback1951OnIA}.
\begin{equation}
    \begin{aligned}
        L_\text{FP} = \sum_{i\neq j} \bm{S}_{ij}^{\text{*}} \log \frac{\bm{S}_{ij}^l}{\bm{S}_{ij}^{\text{*}}},  \bm{S}_{ij}^{\text{*}} = \kappa^\text{Ga}\left(d_{v_i, v_j}^\text{input}, \sigma_i^* \right)
    \end{aligned}
\end{equation}
where $\bm{S}_{ij}^{\text{*}}$ is normalized pairwise similarity in input data, and $d_{v_i, v_j}^\text{input}$ is the distance of $i$ and $j$ in input space. The $\bm{S}_{ij}^l$ is the similarity in the latent space. In t-SNE, $\bm{S}_{ij}^l$ is calculated from the t-distribution. After completing the optimization, the points in latent space are output as visualization results.
}
\section{Methods}
\begin{figure*}[thbp]
    \centering
    \includegraphics[width=7.0in]{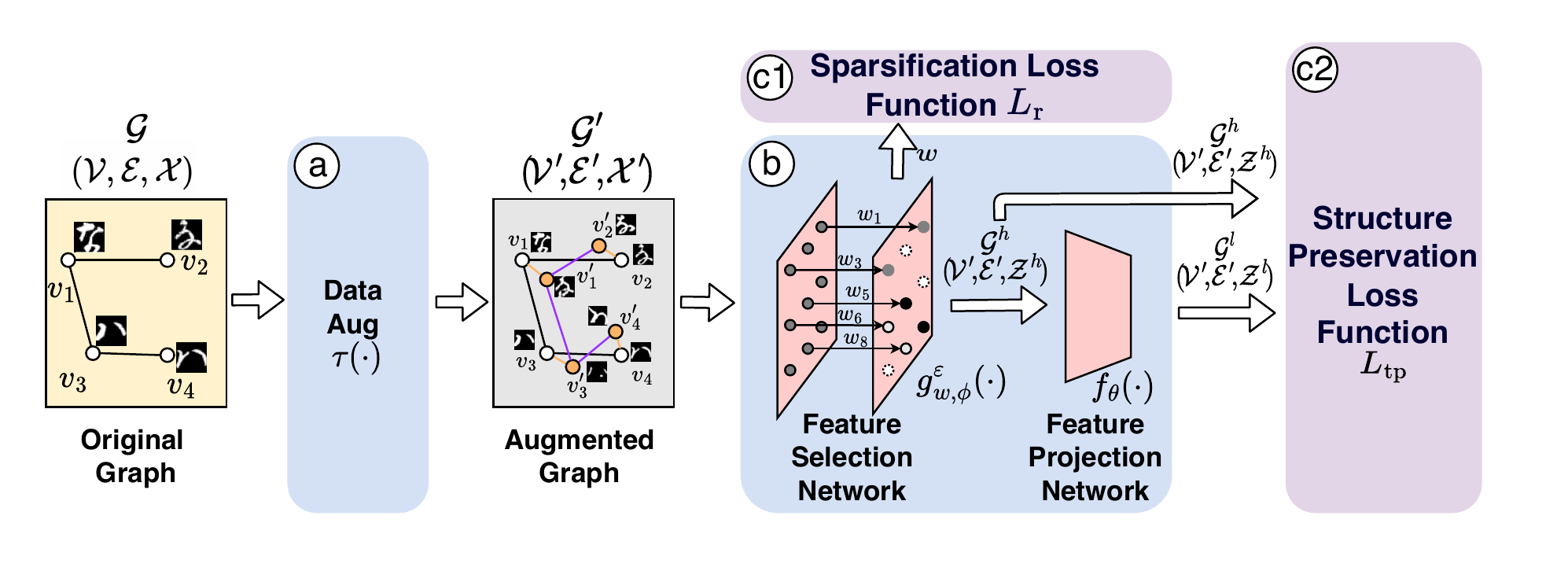}
    % \vspace{-8pt}
    \caption{
        \textbf{The framework of UDRN.}
        The proposed UDRN consists of (a) data augmentation operation $\tau $,  (b) feature selection~(FS) network \& feature projection~(FP) network, (c) structure preservation loss function $L_\text{tp}$.
        The $\tau $ generates new augmented graph $\bm{\mathcal{G'}}(\bm{\mathcal{V'}}, \bm{\mathcal{E'}}, \bm{\mathcal{X'}})$ by Eq.~(\ref{eq_aug_all}).
        The FS network filter out unimportant features and map the data with selected features to high-dim embedding graph $\bm{\mathcal{G}^h}(\bm{\mathcal{V}'}, \bm{\mathcal{E}'}, \bm{\mathcal{Z}^h})$; FP network map the $\bm{\mathcal{G}^h}$ low-dim embedding graph $\bm{\mathcal{G}^l}(\bm{\mathcal{V}'}, \bm{\mathcal{E}'}, \bm{\mathcal{Z}^l})$.
        The $L_\text{tp}$ train FS and FP network with an end-to-end loss function.
    }
    % \vspace{-8pt}
    \label{fig_introduction}
\end{figure*}

% We aim to develop an end-to-end FS\&FP method that leverages the beneficial properties of neural networks and data augmentation to improve the traditional pipeline~(in Fig.~\ref{fig_introduction}).
% The proposed end-to-end neural network is designed to solve the FS\&FP task. 

The network framework of UDRN, the new manifold assumption and the proposed loss function are described in detail in this section. Moreover, the reasons for the performance improvement it brings are further analyzed.

\subsection{UDRN Framework}

As discussed in sec~3.2 to sec~3.4, current FS and FP methods are unable to meet the requirements of FS\&FP task. Therefore, we propose a novel neural network framework to solve the FS\&FP task. The framework of UDRN is shown in Fig.~\ref{fig_introduction}. The proposed UDRN contains a feature selection~(FS) network $g_{\bm{w},\phi}^{\epsilon}(\cdot)$ and a feature projection~(FP) network $f_{\bm{\phi}}(\cdot)$, each oriented to separate aim. The $g_{\bm{w},\phi}^{\epsilon}(\cdot)$ learn the sparse feature~subset \emph{online} and then map the data with selected features into high dimensional embedding $\bm{\mathcal{Z}}^h$, and then $f_{\bm{\phi}}(\cdot)$ further maps $\bm{\mathcal{Z}}^h$ to low dimensional embedding $\bm{\mathcal{Z}}^l$.

The forward propagation of feature selection~(FS) network $g_{\bm{w},\phi}^{\epsilon}(\cdot)$ is,
\begin{equation}
    \begin{aligned}
        \bm{\mathcal{Z}}^h & = g_{\bm{w},\phi}^\epsilon(\bm{\mathcal{X'}}) =
        % \bm{\mathcal{W}} \circ \bm{1}_{\bm{\mathcal{W}} >\epsilon} 
        \text{m}_\phi \left( \text{GL}_{\bm{w}}^{\epsilon} \left( \bm{\mathcal{X'}} \right) \right), %g_{\bm{w},\phi}^\epsilon(\bm{\mathcal{X'}})
        % \\
        % \bm{\mathcal{Z}}^h & = f_{\bm{\theta}} (\bm{\mathcal{X}}^h)
    \end{aligned}
    \label{equ_fsnet}
\end{equation}
{\color{black}
where the FS network includes a backbone network $\text{m}_{\phi} \left( \cdot \right)$ and a gate layer $\text{GL}_{\bm{w}}^{\epsilon}\left( \cdot \right)$.
The $\phi$ is the network parameters of $\text{m}_{\phi} \left( \cdot \right)$.
The gate layer $\text{GL}_{\bm{w}}^{\epsilon}\left( \cdot \right)$ processes the augmented data $\bm{\mathcal{X'}}$ by a gate operation.
\begin{equation}
    \begin{aligned}
        \text{GL}_{\bm{w}}^{\epsilon}(\bm{\mathcal{X'}}) = \left\{
        \begin{array}{cl}
            w_j \bm{\mathcal{X}_j '} & \text{if } w_j > \epsilon, j \in \left\{ 1, 2, \ldots, \mathcal{D}\right\} \\
            0                             & \text{otherwise}
        \end{array}
        \right.
    \end{aligned}
    \label{gatelayer}
\end{equation}
where the $\bm{w} \in R^{\mathcal{D}}$ is gate parameters, indicating the importance of the features. The $\epsilon$ is a hyperparameter threshold, and the $g_{\bm{w},\phi}^\epsilon(\bm{\mathcal{X'}})$ is a gate layer to ensure the features with low importance scores are not leakage to the latter network layer.

In this way, important features~($w_j > \epsilon$) can be passed through the gate layer and scaled by the gate parameters. And unimportant features will be blocked by the gate layer.  The $\bm{w}$ is initialized to a constant value and is optimized according to the loss function of the network. Once $w_j < \epsilon$, the corresponding feature $j$ is discarded by the gate layer.
}

The forward propagation of the feature propagation~(FP) network is,
\begin{equation}
    \begin{aligned}
        \bm{\mathcal{Z}}^l = f_{\bm{\theta}} (\bm{\mathcal{Z}}^h).
    \end{aligned}
    \label{equ_fpnet}
\end{equation}
The FP network $f_{\bm{\theta}}(\cdot)$ maps the h-dim embedding $\bm{\mathcal{Z}}^h$ to the $\bm{\mathcal{Z}}^l$ in l-dim for visulazation or other downstreem tasks.
Finally, the outputs of UDRN are two graphs, which include the high dimensional embedding graph $\bm{\mathcal{G}}^h(\bm{\mathcal{V'}}, \bm{\mathcal{E'}}, \bm{\mathcal{Z}^h})$ and low dimensional embedding graph $\bm{\mathcal{G}}^l(\bm{\mathcal{V'}}, \bm{\mathcal{E'}}, \bm{\mathcal{Z}^l})$.

    {
        \color{black}
        \subsection{Manifold Connectivity Assumptions Under Augmented Data}

        Typical FS and FP methods are based on manifold assumptions~\cite{lin2008riemannian, agarwal2010learning, kobak_umap_2019, MarziehEdraki2021ImplicationOM}, which only focus on finite data in the dataset. These methods ignore estimating the intrinsic manifold from the finite and augmented data. It is a good choice to learn information about the intrinsic manifold by neural networks and to perform FS and FP based on the learned network parameters containing the information of the manifold.

        To realize the above plan, the current manifold assumption has to be extended as it is not compatible with augmented data. As shown in Fig.~\ref{fig_augment}, the typical DR loss does not take into account the augmented data, so the embedding of the latent space can only be learned based on the finite data. The `push' and `pull' are the figurative representation of the action of the loss function on the nodes of the latent space. Current typical methods cannot accommodate data augmentation. When additional augmented data is generated, the loss function of the current typical method causes an increase in computational complexity~(because of the need to compute $\sigma^{*}$ for each augmented data) and model collapse (because of the inconsistency of the gradient caused by the data augmentation).

        \begin{figure}[h]
            \centering
            \color{black}
            \includegraphics[width=3.2in]{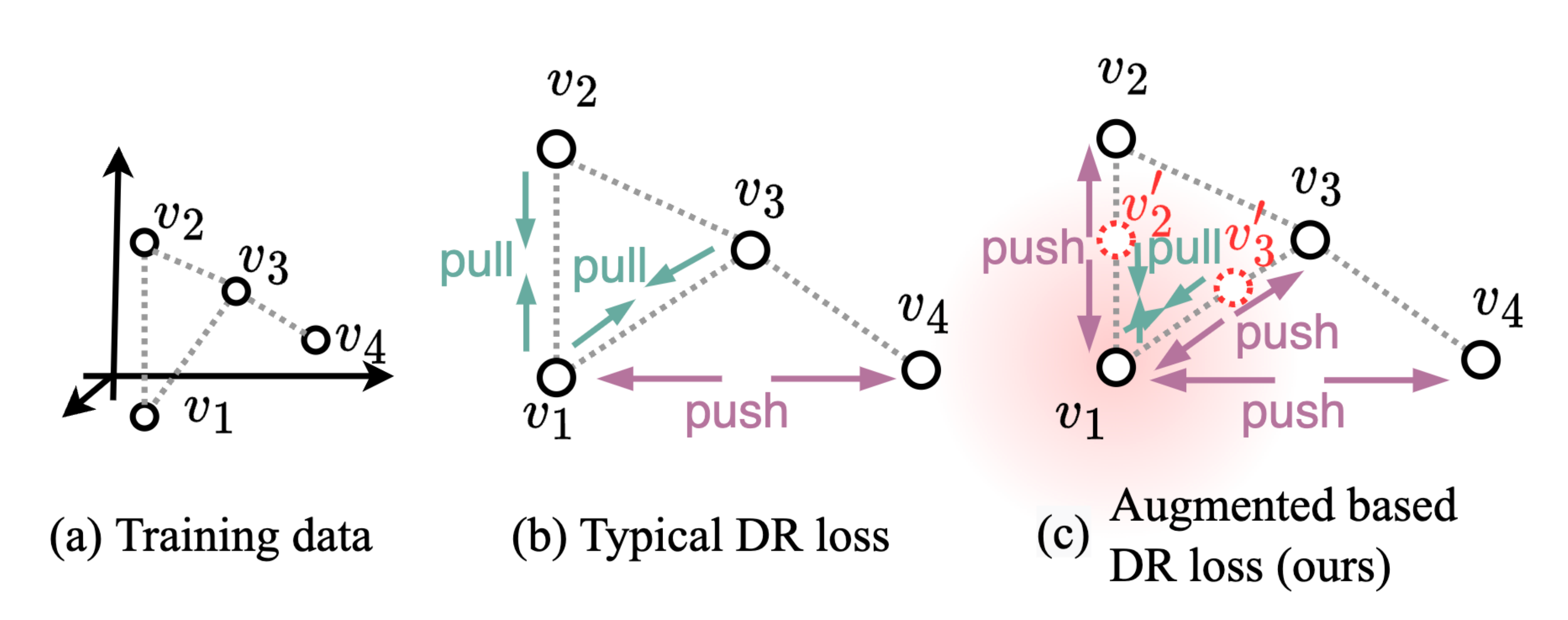}
            \caption{\textbf{Manifold connectivity assumptions in data augmentation contexts.} (a) High dimensional data used for training model. (b) Typical DR loss by `pushing and pulling' the true sample to optimize the latent space. (c) The proposed DR method based on data augmentation generates new samples online and trains the model by `pushing and pulling' augmented samples to learn the intrinsic manifold more precisely.
            }
            \label{fig_augment}
        \end{figure}

        To this end, we propose a more stringent assumption, named manifold connectivity assumptions under augmented data. It assumes that the augmented data $\bm{x}^{'}=\tau(\bm{x})$ is connected to the original data $x$ on the manifold. Based on this, the typical DR loss can be effectively expanded~(as shown in Fig.~\ref{fig_augment}).
        Instead of relying on the finite data in the dataset to optimize the neural network parameters, the proposed loss is combined with the augmented data to train the model. 
        A sufficient prior embedded in the data augmentation allows the model to be trained without computing $\sigma^{*}$, which also avoids additional computational consumption. Also, the proposed assumptions can better avoid collapse and achieve more refined modeling.
    }

\subsection{Data Augmentation Compatible Loss Functions}

{\color{black}
Next, a novel loss function is designed to implement the assumptions proposed in Sec~4.2. The proposed loss function matches the network framework~(in Sec~4.1) and data augmentation~(Sec~3.3).

% We find that no current technique can address end-to-end training of neural networks for FS and FP tasks. 
% Typical FS approaches are often based on reconstruction losses which can not guarantee training out a latent space where the neighbor structure of the input data is preserved. At the same time, typical FP methods cannot select important data features. It is also worth noting that simply combining FP and FS tasks into a neural network does not guarantee that the network parameters are well-optimized. To address the problem of better optimization of network parameters, we introduce data augmentation and manifold connectivity assumptions to design a novel loss function.

The proposed loss function first ensures the preservation of local information in the dimensionality reduction process by \textit{manifold exaggeration} and then measures the difference between the latent space and the target with the \textit{fuzzy set cross entropy}~(in Fig.~\ref{fig_SPloss}).

\begin{figure}[h]
    \centering
    \includegraphics[width=3.2in]{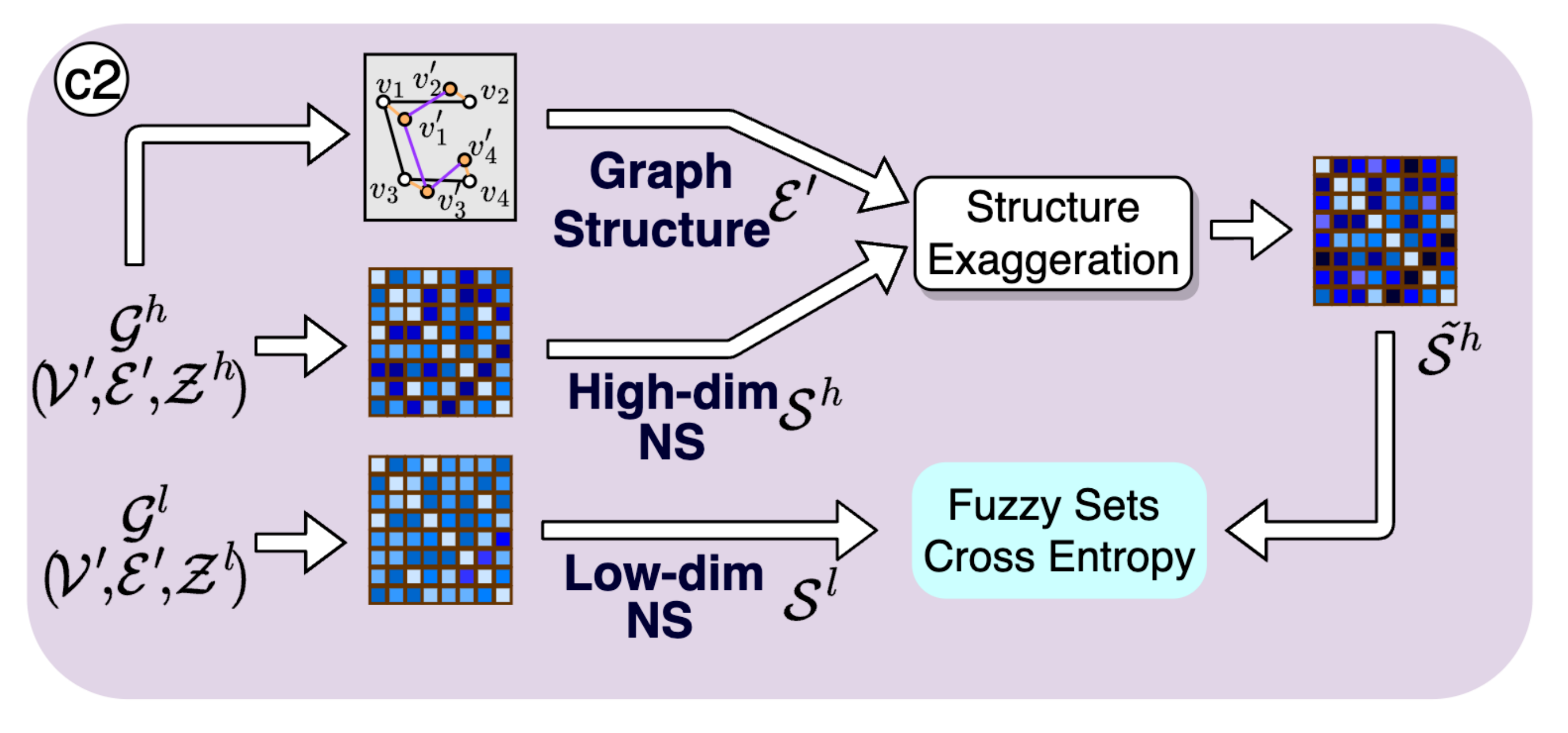}
    \caption{
        Framework of calculating $L_{\text{tp}}$.
        (1) abstract structure $\bm{\mathcal{E'}}$.
        (2) use $\bm{\mathcal{E'}}$ to exaggerate the high-dimensional NS $\mathcal{\bm{S}}^{\bm h}$.
        (3) calculate $L_{\text{tp}}$ by appling fuzzy set cross-entropy to $\mathcal{\bm{\tilde{S}}}^{\bm h}$ and $\mathcal{\bm{S}}^{\bm l}$.
    }
    \label{fig_SPloss}
\end{figure}

The DR requires mapping high-dimensional data to a lower-dimensional space, which naturally brings about `crowding problem'. {\color{black} To alleviate the above `crowding problem', pulling in neighboring nodes and pushing away non-neighboring nodes are good strategies. Because the above strategies can effectively avoid the manifold overlapping in the lower dimensional space. Thus we design \textit{manifold exaggeration},}
\begin{equation}
    % \large
    \begin{aligned}
        \tilde{\bm{S}}_{ij}^{h} & = E(\bm{S}_{ij}^{h}, \bm{\mathcal{E'}}) \\
                                & = \left\{
        \begin{aligned}
             & {\bm{S}}_{ij}^{h} \exp(1-\beta) & \text{ if } \left(\bm{v}_{i}, \bm{v}_{j}\right)  \in \bm{\mathcal{E'}} & \\
             & {\bm{S}}_{ij}^{h} \exp(1+\beta) & \text{otherwise}                                                       &
        \end{aligned}
        \right.
    \end{aligned}
\end{equation}
where the neighborhood relationship of augmented data $\bm{\mathcal{E'}}$ is defined in Eq.~(\ref{eq_aug_all}). The $\bm{S}_{ij}^{h}$ is the node similarity, which is calculated from the nodes similarity in a high dimensional graph $\bm{\mathcal{G}^h}(\bm{\mathcal{V}'}, \bm{\mathcal{E}'}, \bm{\mathcal{Z}^h})$.
$
    \bm{S}_{ij}^{h} = \kappa^\text{Ga}\left(\left\|\bm{z}_i^h-\bm{z}_j^h\right\|_{2}^2, \sigma=1 \right)
$.
The \textit{manifold exaggeration} transforms the goal of network learning with neighbor relationship prior knowledge $\bm{\mathcal{E'}}$, that is, increasing the goal similarity of neighboring nodes and decreasing the goal similarity of non-neighboring nodes. Thereby, the objective of pulling in neighbor nodes and pushing away non-neighbor nodes can be achieved. The hyperparameter $\beta>0$ controls the strength of exaggeration.

The loss function is designed as the form of fuzzy sets cross entropy~\cite{kobak_umap_2019} (two-way divergency in~\cite{li_deep_2020}),
\begin{equation}
    \begin{aligned}
        L_\text{tp} & = \frac{1}{B^2}\sum_{i,j \in \bm{\mathcal{B}}}  \tilde{\bm{S}}_{ij}^{h} \log \bm{S}_{ij}^l + (1-\tilde{\bm{S}}_{ij}^{h}) \log (1-\bm{S}_{ij}^l) \\
        % \bm{S}^Y & = \{s_{ij}^Y \ \ |\ \ s_{ij}^Y=e^{-\| y_i - y_j \|^2}, y_i,y_j \in Y\}
    \end{aligned}
    \label{equ_losstp}
\end{equation}
where nodes similarity in a high dimensional graph is calculated as $\bm{S}_{ij}^l= \kappa^\text{t}\left(\left\|\bm{z}_i^l-\bm{z}_j^l\right\|_{2}^2, \nu \right)$, $\nu$ is a hyperparameter. The $\bm{\mathcal{B}}=\{1, 2, \ldots, B\}$, $B$ is the number of node in a batch. The loss function trains the neural network to output a low-dimensional embedding $z^l$ such that $\bm{S}_{ij}^l$ approximates the exaggerated $\tilde{\bm{S}}_{ij}^{h}$.

% The designed manifold exaggeration has a consistent loss function for FS and FP.

The designed loss function is based on the \textit{manifold connectivity assumption of augmented data} and is well-compatible with data augmentation. As described in Sec.~4.3, it is assumed that the augmented data are neighbors of the original data in the real manifold. Therefore, instead of pulling in similar nodes in the dataset, the proposed loss function pulls in the augmented nodes. The online generation of augmented data during training allows the proposed method to depict the structure of the manifold in a more refined way, ultimately leading to performance improvements.

The designed loss function is consistent for FS and FP. We implement the selection of important features and the discarding of unimportant features in the forward propagation of the network with the help of the gate layer. And the whole selection process is embedded in the training of the neural network. The FS and FP are based on the same object, which is to better preserve the structure of neighbors in higher dimensional spaces in a lower dimensional space.
}

Finally, the loss function of unsupervised UDRN is\@:
\begin{equation}
    \begin{aligned}
        \min _{w, \bm{\theta}, {\bm{\phi}}} \ \
        L_\text{tp} + \lambda L_\text{r}, \ \ \ \ L_\text{r}=\|\bm{w}\|_1 , \\
    \end{aligned}
    \label{eq:final_model}
\end{equation}
where $\lambda$ is hyperparameter. To select a specific feature number, we give $\lambda$ a small initial value, and then slowly increase $\lambda$ until the feature number satisfies the requirements.

\subsection{Pseudocode}
Algorithm.~\ref{alg_algorithm} shows how to train our model and how to obtain the selected features.

\begin{algorithm}[tpb]
    \caption{The UDRN algorithm}
    % \small
    \label{alg_algorithm}
    \textbf{Input}: 
        Data: $\bm{\mathcal{G}_{\text{all}}}(\bm{\mathcal{V}}, \bm{\mathcal{E}}, \bm{\mathcal{X}})$, 
        Learning rate: $\alpha$, 
        Epochs: $E$,
        Batch size: $B$,
        Network: $g_{w,\theta}^{\epsilon}, f_{\theta}$,
        loss weights: $\lambda$,
        \\
    \textbf{Output}: 
    Selected Features: $\bm{\mathcal{X}}^h$.
    FS\&FP Embedding: $\bm{\mathcal{Z}_\text{all}^l}$.\\
    
    \begin{algorithmic}[1] %[1] enables line numbers
        \STATE Let $t=0$.
        \WHILE{{$i=0$; $i<E$; $i$++}}
            \WHILE{{$b=0$; $b<[ |\mathcal X| /B]$; $b$++}}
            % \STATE $\mathcal{B} = \{1,\cdots, B\}$
            \STATE $\bm{\mathcal{G}}(\bm{\mathcal{V}}, \bm{\mathcal{E}}, \bm{\mathcal{X}})$ 
            $\leftarrow$
            Sampling($\bm{\mathcal{G}_{\text{all}}}(\bm{\mathcal{V}}, \bm{\mathcal{E}}, \bm{\mathcal{X}})$, b); \\ 
            {\color{OliveGreen} \# Sample a batch data} \\
            \STATE 
            $\bm{\mathcal{G}}'(\bm{\mathcal{V}}', \bm{\mathcal{E}}', \bm{\mathcal{X}}')$ 
            $\leftarrow$ 
            Augment($\bm{\mathcal{G}}(\bm{\mathcal{V}}, \bm{\mathcal{E}}, \bm{\mathcal{X}})$) by Eq.~(\ref{eq_aug_all});\\ 
            {\color{OliveGreen} \# Data augmentation} \\
            \STATE 
            $\bm{\mathcal{X}}^h = m(\bm{\mathcal{X'}})$ by Eq.~(\ref{equ_fsnet}); \\
            {\color{OliveGreen} \# Select the features}\\
            \STATE 
            $\bm{\mathcal{G}^h}(\bm{\mathcal{V}'}, \bm{\mathcal{E}'}, \bm{\mathcal{Z}^h})$ 
            $\leftarrow$ 
            $g(\bm{\mathcal{G'}}(\bm{\mathcal{V'}}, \bm{\mathcal{E'}}, \bm{\mathcal{X}}^h))$ by Eq.~(\ref{equ_fsnet});\\
            {\color{OliveGreen} \# Map to high dimension space}\\
            \STATE 
            $\bm{\mathcal{G}^l}(\bm{\mathcal{V}'}, \bm{\mathcal{E}'}, \bm{\mathcal{Z}^l})$ 
            $\leftarrow$ 
            $f(\bm{\mathcal{G}^h}(\bm{\mathcal{V}'}, \bm{\mathcal{E}'}, \bm{\mathcal{Z}^h}))$ by Eq.~(\ref{equ_fpnet});\\
            {\color{OliveGreen} \# Map to low dimension space}\\
            \STATE 
            $\bm{S}_{ij}^{h}$ 
            $\leftarrow$ 
            $S(\bm{\mathcal{G}^h}(\bm{\mathcal{V}'}, \bm{\mathcal{E}'}, \bm{\mathcal{Z}^h}))$;
            {\color{OliveGreen} \# Cal high dim similarity}\\
            \STATE 
            $\bm{S}_{ij}^{l}$ 
            $\leftarrow$ 
            $S(\bm{\mathcal{G}^l}(\bm{\mathcal{V}'}, \bm{\mathcal{E}'}, \bm{\mathcal{Z}^l}))$;
            {\color{OliveGreen} \# Cal low dim similarity}\\
            \STATE 
            $\tilde{\bm{S}}_{ij}^{h}$ 
            $\leftarrow$ 
            $E(\bm{S}_{ij}^{h}, \bm{\mathcal{E'}})$;
            % $  $ 
            {\color{OliveGreen} \# Manifold exaggeration}\\
            \STATE $\mathcal{L}_{\text{tp}} \!\leftarrow\! L_\text{tp}(\tilde{\bm{S}}_{ij}^{h}, \bm{S}_{ij}^{l})$ by Eq.~(\ref{equ_losstp});
            {\color{OliveGreen} \# Cal. the structural preservation loss} \\
            \STATE $\mathcal{L}_{\text{1}} \!\leftarrow\! L_\text{1}(\bm{w})$ by Eq.~(\ref{eq:final_model});
            {\color{OliveGreen} \# Cal. the L1 loss} \\
            \STATE $\theta,\!\leftarrow\!  \theta- \alpha 
            \frac{ \partial \mathcal{L}_\text{tp} }{\partial \theta}$, 
            $\phi \!\leftarrow\!  \phi - \alpha 
            \frac{ \partial \mathcal{L}_\text{tp} }{\partial \phi}$, 
            $\bm{w} \!\leftarrow\! \bm{w} - \alpha 
            (
                \frac{ \partial \mathcal{L}_\text{tp} }{\partial \bm{w}}+\frac{ \partial \mathcal{L}_\text{1} }{\partial \bm{w}}
            )$
            {\color{OliveGreen} \# Update the parameters} \\
            \ENDWHILE
        \ENDWHILE
        \STATE $\bm{\mathcal{X}}^h \leftarrow m(\bm{\mathcal{G}_{\text{all}}}(\bm{\mathcal{V}}, \bm{\mathcal{E}}, \bm{\mathcal{X}}))$;
        {\color{OliveGreen} \# Select the features} \\
        \STATE $\bm{\mathcal{G}^l_\text{all}}(\bm{\mathcal{V}_\text{all}'}, \bm{\mathcal{E}_\text{all}'}, \bm{\mathcal{Z}_\text{all}^l}) \leftarrow f(g(\bm{\mathcal{G}_{\text{all}}}(\bm{\mathcal{V}}, \bm{\mathcal{E}}, \bm{\mathcal{X}})))$;\\
        {\color{OliveGreen} \# Cal. the embedding result} \\
    \end{algorithmic}
\end{algorithm}
\section{Experiments}

\subsection{Details of Dataset and Compared Methods}
\label{app_sec_Dataset}
\textbf{Details of Dataset.}
We used four image datasets (COIL20, Mnist, KMnist, EMnist) and four biological datasets~(Activity, HCL, Gast, and MCA). {\color{black} The details of the dataset are shown in Table~\ref{tab_dataset}}.
Unlike CAE~\cite{Abubakar} and FAE~\cite{XinxingWu2021FractalAF}, we do not downsample the dataset because of the computational time. We consider performance on large datasets as an essential evaluation metric.
\begin{table*}[thbp]
    \caption{Statistics of datasets. }
    \small
    \centering
    \begin{tabular}{c|ccccm{7cm}}
        \toprule
                                                                                 & Dataset  & \#Sample & \#Feature & \#Class & Link                                                                     \\ \midrule
        \multirow{4}{*}{\begin{tabular}[c]{@{}c@{}}Image Data\end{tabular}}      & Coil20   & 1440     & 16384     & 20      & https://www.cs.columbia.edu/CAVE/ software/softlib/coil-20.php           \\
                                                                                 & KMnist   & 60,000   & 784       & 10      & https://pytorch.org/vision/stable/index.html                             \\
                                                                                 & Mnist    & 60,000   & 784       & 10      & https://pytorch.org/vision/stable/index.html                             \\
                                                                                 & EMnist   & 731,668  & 784       & 10      & https://pytorch.org/vision/stable/index.html                             \\ \midrule
        % \multirow{6}{*}{\begin{tabular}[c]{@{}c@{}}Biological Data\end{tabular}} & pixraw10P  & 100        & 10000     & 10      & https://jundongl.github.io/scikit-feature/datasets.html                  \\
        %                                                                          & Prostatege & 102        & 5966      & 2       & https://sites.google.com/site/feipingnie/file/                           \\
        \multirow{4}{*}{\begin{tabular}[c]{@{}c@{}}Biological Data\end{tabular}} & Activity & 5,744    & 561       & 6       & https://www.kaggle.com/uciml/human-activity-recognition-with-smartphones \\
                                                                                 & GAST     & 10,629   & 1457      & 12      & https://www.ncbi.nlm.nih.gov/pmc/articles/PMC4643992/                    \\
                                                                                 & MCA      & 30,000   & 9119      & 52      & http://bis.zju.edu.cn/MCA/                                               \\
                                                                                 & HCL      & 280,000  & 3037      & 93      & https://figshare.com/articles/dataset/ HCL\_DGE\_Data/7235471            \\

        \bottomrule
    \end{tabular}
    \label{tab_dataset}
\end{table*}

\textbf{Compared Methods.}
To demonstrate the advantages of UDRN, we compare it with the FS methods, the FP methods, and the pipeline methods of FS and FP\@.
The compared methods are divided into non-parameters methods and parameters methods.
The non-parameters FS methods include
LS~{\cite{he2006laplacian}},
MCFS~{\cite{cai_unsupervised_2010}},
NDFS~{\cite{li_unsupervised_2012}},
and IVFS~{\cite{li2020ivfs}}.
The compared parameters FS methods include
AEFS~{\cite{Han}},
CAE~{\cite{Abubakar}},
FAE~{\cite{wu_fractal_2020}},
and QS~{\cite{atashgahi2021quick}}.
The compared non-parameters FP methods include
tSNE~\cite{kobak2019art},
UMAP~\cite{mcinnes2018umap}.
The compared parameters FP methods include
GRAE~\cite{duque2020extendable},
IVIS~\cite{Szubert_Cole_Monaco_Drozdov_2019} and
Parametric UMAP~(PUMAP)~\cite{sainburg2021parametric}.

The grid search is used to determine the optimal parameters for all the baseline methods. The search space of each method is shown in Table.~\ref{tab_DetailsofGridSearch}.

% In sec.~\ref{sec_Comparison_with_FS_methods}, we use the grid search method to determine the optimal hyper-parameters.
% The details are in Table.~\ref{tab_DetailsofGridSearch}.

\begin{table*}[tbp]
    \small
    % \centering
    \caption{Details of grid search.}
    % \vspace{-8pt}
    \begin{tabular}{l|c|c}
        \toprule
        Methods & Search Space                                             & Note                                                 \\ \midrule
        LS      & $C\in[5,10,15,20,25,30,35,40,45,50]$                     & $C \to \text{cluster range}$                                \\
        MCFS    & $C\in[5,10,15,20,25,30,35,40,45,50]$                     & $C \to \text{cluster range}$                                \\
        NDFS    & $A\in[1,1.5,2]$,$BETA\in[0.5,1,2]$,$C\in[5,15,25]$       & $A \to \alpha$, $C \to \text{cluster range}$                 \\
        IVFS    & $T\in[D//10,D//20,D//50,D//80]$                          & $T \to \text{tilde sample range},D \to \text{data shape}$          \\
        AEFS    & $A\in[0.1,0.2,0.5]$, $E\in[500,1000,2000]$               & $A \to \alpha, E \to \text{epoch}$                           \\
        CAE     & $B\in[256,512]$, $LR\in[0.01,0.1,1]$, $DR\in[0,0.5,0.8]$ & $B \to \text{batch}, LR \to \text{learning rate} , DR \to \text{dropout}$ \\
        % PFA     & $q\in[0.1,0.3,0.5,0.7,0.9] $                                                                                    \\
        FAE     & $B\in[128,256,512]$,$E\in[500,1000,2000]$                & $B \to \text{batch}$, $E \to \text{epoch}$                         \\
        QS      & $EP\in[2,5,10,13,20,25]$,$Z\in[0.1,0.2,0.3,0.4,0.5]$     & $EP \to \epsilon$, $Z \to \zeta$                       \\
        \bottomrule
    \end{tabular}
    % \vspace{-8pt}
    \label{tab_DetailsofGridSearch}
\end{table*}

\subsection{Experimental Setup}

We initialize the weights of the FS layer to $\bm{w}=0.2$ and initialize the other NN with the Kaiming initializer.
We adopt the AdamW optimizer~\cite{loshchilov2017decoupled} with a learning rate of 0.001.
All experiments use a fixed MLP network structure, $g_{\bm{w}}^{\epsilon}$: [-1, 500, 300, 80], $f_{\theta}$: [80, 500, 2], where -1 is the features number of the dataset, the first layer of $g_{\bm{w}}^{\epsilon}$ is the gate layer.
To make UDRN select a specified number of features, we set an adaptive $\lambda$. At the beginning of 300 epochs, the $\lambda=0$ model, and then $\lambda= L_\text{r} / 0.1 \|\bm{w}\|_1$ and grow by 0.5\% until the number of features meet the requirements. For all experiments $\beta=0.01$.
For the experiments in Table~\ref{tab_class_acc_image} to Table~\ref{tab_fp}, we used Bernoulli-type FMH augmentation and set $p_N = 0.3$. For a fair comparison, the training set~(80\% data) is used for the model training and feature selection; the validation set~(10\%) is used to select the best hyperparameters with grid search; the performance on the test set~(10\%) is reported in this paper.
\begin{table*}[t]
    \centering
    \caption{Discriminative performance~(classification accuracy) comparison with FS methods in image datasets; best result are shown in \textbf{blod}; results with clear advantage are shown in \underline{underline}.}
    % \vspace{-8pt}
    \small
    \label{tab_class_acc_image}
    \begin{tabular}{l|cccc|cccc|c}
        \toprule
        {}                   & LS           & MCFS         & NDFS         & IVFS         & AEFS         & CAE          & FAE          & QS           & UDRN                                             \\ \midrule
        % {}                      & \tiny{\cite{he2006laplacian}} & \tiny{\cite{cai_unsupervised_2010}} & \tiny{\cite{li_unsupervised_2012}} & \tiny{\cite{li2020ivfs}} & \tiny{\cite{Han}}      & \tiny{\cite{Abubakar}} & \tiny{\cite{wu_fractal_2020}} & \tiny{\cite{atashgahi2021quick}} & (ours)                \\ \midrule
        {Coil20}  & 21.0$\pm$0.6 & 34.0$\pm$1.3 & 8.1$\pm$1.5  & 98.6$\pm$0.7 & 99.3$\pm$0.2 & 97.7$\pm$0.7 & 84.1$\pm$0.2 & 98.0$\pm$0.5 & \textbf{99.4$\pm$0.2($\uparrow$0.1)}             \\
        {MNIST}   & 17.0$\pm$0.1 & 76.0$\pm$0.4 & 90.4$\pm$0.6 & 42.4$\pm$0.1 & 86.4$\pm$0.3 & 92.1$\pm$0.2 & 70.5$\pm$0.4 & 93.2$\pm$0.2 & \textbf{94.3$\pm$0.3($\uparrow$1.1)}             \\
        {KMNIST}  & 20.1$\pm$0.2 & 64.0$\pm$0.5 & 83.9$\pm$0.3 & 82.4$\pm$0.5 & 85.6$\pm$0.4 & 88.0$\pm$0.3 & 77.6$\pm$0.2 & 85.9$\pm$0.3 & \underline{\textbf{90.7$\pm$0.4($\uparrow$2.7)}} \\
        {EMNIST}  & 7.9$\pm$0.1  & 43.6$\pm$0.9 & 64.3$\pm$0.6 & 42.5$\pm$0.3 & 65.6$\pm$0.4 & 63.9$\pm$0.3 & 52.0$\pm$0.3 & 68.0$\pm$0.3 & \underline{\textbf{71.1$\pm$0.5($\uparrow$3.1)}} \\ \midrule
        {Average} & 16.5$\pm$0.2 & 54.4$\pm$0.6 & 61.7$\pm$0.8 & 66.5$\pm$0.5 & 84.2$\pm$0.3 & 85.4$\pm$0.4 & 64.8$\pm$0.3 & 86.2$\pm$0.3 & \underline{\textbf{88.9$\pm$0.4($\uparrow$2.7)}} \\
        \bottomrule
    \end{tabular}
\end{table*}

\begin{table*}[t]
    \centering
    \caption{Discriminative performance~(classification accuracy) comparison with FS methods in biology datasets, best result are shown in \textbf{blod}; results with clear advantage are shown in \underline{underline}. }
    % \vspace{-8pt}
    \small
    \label{tab_class_acc_bio}
    \begin{tabular}{l|cccc|cccc|c}
        \toprule
        {}                    & LS           & MCFS         & NDFS         & IVFS         & AEFS         & CAE          & FAE          & QS           & UDRN                                              \\ \midrule
        % {}                      & \tiny{\cite{he2006laplacian}} & \tiny{\cite{cai_unsupervised_2010}} & \tiny{\cite{li_unsupervised_2012}} & \tiny{\cite{li2020ivfs}} & \tiny{\cite{Han}}      & \tiny{\cite{Abubakar}} & \tiny{\cite{wu_fractal_2020}} & \tiny{\cite{atashgahi2021quick}} & (ours)                \\ \midrule
        % {pixraw10P}  & 89.0$\pm$3.0                  & 60.0$\pm$9.1                        & 20.0$\pm$0.1                       & \textbf{100$\pm$0.1}     & \textbf{100.0$\pm$0.1} & \textbf{100$\pm$0.1}   & 98.0$\pm$8.0                  & \textbf{100$\pm$0.1}             & \textbf{100$\pm$0.1}  \\
        % {Prostatege} & 74.0$\pm$6.6                  & 46.0$\pm$9.1                        & 58.0$\pm$6.0                       & 80.0$\pm$4.4             & 89.0$\pm$3.0           & 90.0$\pm$0.1           & 74.0$\pm$4.8                  & 86.0$\pm$4.8                     & \textbf{90.0$\pm$0.1} \\
        {Activity} & 92.3$\pm$0.3 & 42.4$\pm$1.8 & 48.5$\pm$3.8 & 95.8$\pm$0.3 & 96.9$\pm$0.2 & 98.0$\pm$0.1 & 74.8$\pm$0.8 & 97.4$\pm$0.2 & {\textbf{98.6$\pm$0.3($\uparrow$0.6)}}            \\
        {HCL}      & 23.6$\pm$0.2 & 7.2 $\pm$0.5 & 09.2$\pm$0.2 & 21.8$\pm$0.2 & 24.7$\pm$0.2 & 28.5$\pm$0.1 & 33.0$\pm$0.2 & 56.7$\pm$0.3 & \underline{\textbf{58.9$\pm$0.2($\uparrow$2.2)}}  \\
        {Gast}     & 68.9$\pm$0.5 & 42.1$\pm$3.6 & 44.4$\pm$1.6 & 73.9$\pm$0.6 & 73.6$\pm$0.6 & 89.1$\pm$0.3 & 81.0$\pm$0.3 & 86.8$\pm$0.6 & {\textbf{90.0$\pm$0.4($\uparrow$0.9)}}            \\
        {MCA}      & 19.5$\pm$0.2 & 24.5$\pm$0.1 & 56.8$\pm$0.6 & 27.0$\pm$0.1 & 28.6$\pm$0.2 & 66.8$\pm$0.2 & 32.8$\pm$0.2 & 64.3$\pm$0.4 & \underline{\textbf{77.8$\pm$0.4($\uparrow$11.0)}} \\ \midrule
        {Average}  & 42.4$\pm$0.3 & 31.9$\pm$1.2 & 44.6$\pm$1.4 & 52.2$\pm$0.3 & 57.8$\pm$0.3 & 69.2$\pm$0.2 & 54.3$\pm$0.4 & 74.6$\pm$0.3 & \underline{\textbf{81.3$\pm$0.3($\uparrow$6.7)}}  \\
        \bottomrule
    \end{tabular}
    % \vspace{-8pt}
\end{table*}

\subsection{Case Study}
\begin{figure*}[t]
    \centering
    \includegraphics[width=6.5in]{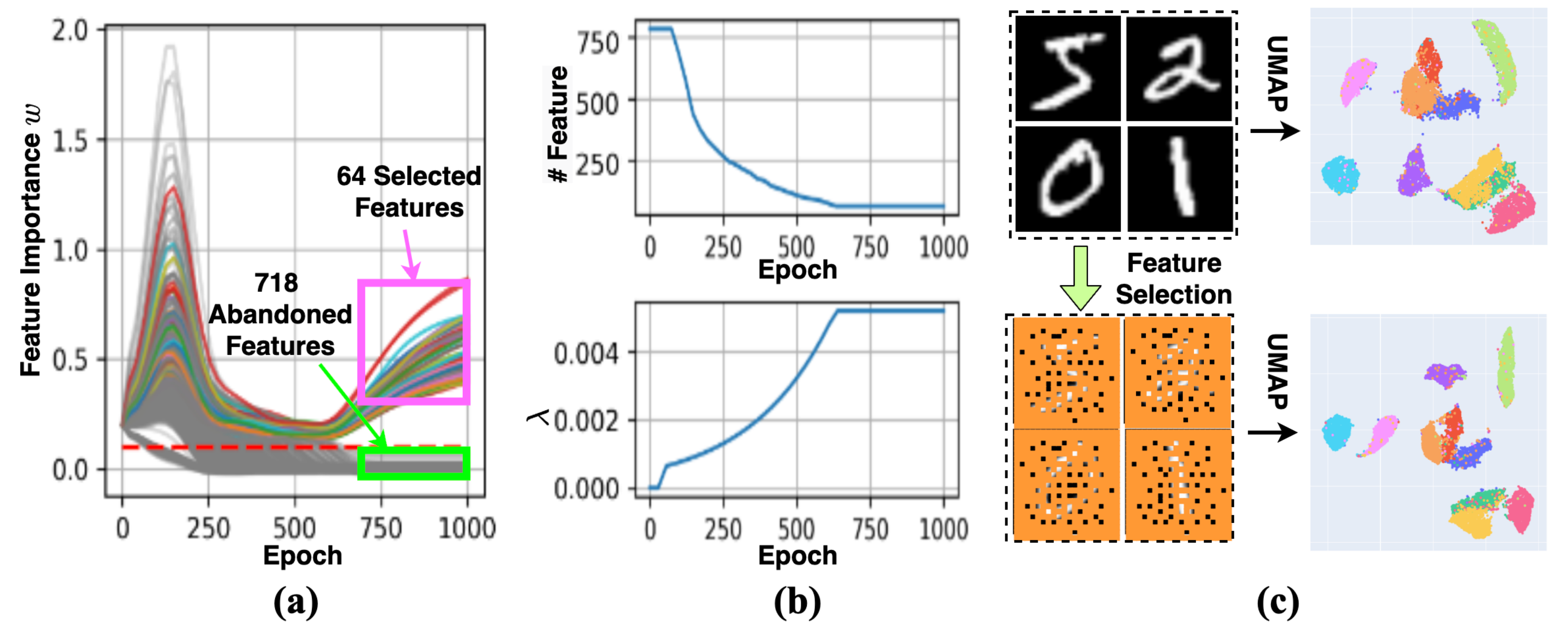}
    % \vspace{-8pt}
    \caption{
        Case Study.
        (a) Variation plot of feature importance during training.
        Curves show the change in feature importance.
        colored curves~(final selected features);
        gray curves~(final discarded features).
        (b) Variation plot of the number of essential features ($w_f > \epsilon$) and loss function weights $\lambda$ during training.
        (c) UMAP Visualization of all features and the selected features. { \color{black} The color of the points in the scatter plot marks the true label of the data.}
    }
    % \vspace{-8pt}
    \label{fig_case_study}
\end{figure*}
The Mnist dataset is selected to illustrate how UDRN works~(Fig.~\ref{fig_case_study}).
\textbf{The FS processing.}
{\color{black}
    At the beginning of training, we set $\bm{w}=0.2$. The gate layer passes all features in the dataset. With the training going on, the $L_\text{r}$ loss reduces $\bm{w}$ and $L_\text{tp}$ loss increases $\bm{w}$. Eventually, only features that are important for structure preservation can pass the gate layer. The unimportant features are discarded~(as shown in Fig.~\ref{fig_case_study}~(a) and Fig.~\ref{fig_case_study}~(b)).}

\textbf{FS \& structure-preservation.}
We expect the FS of UDRN to affect the local and global structure of the data as little as possible, which is the original intention of using the unified loss function for both FS and FP tasks.
We find that humans can easily recognize numbers in images based on selected features, indicating that our FS does not destroy the discriminative nature of the images~(in Fig.~\ref{fig_case_study}(c)). To further confirm this, UMAP is used to process the pre-FS data and post-FS data. The results~(in Fig.~\ref{fig_case_study}(c)) show that the clustering relationship of the data is found not to be changed by FS\@.

\subsection{Comparison with FS methods}
\label{sec_Comparison_with_FS_methods}
This sub-section compares the performance with the unsupervised FS methods. The performance comparison includes discriminative performance and structure-preservation performance.

\textbf{Discriminative performance.}
The discriminative performance shows the ability of the features selected by the FS method in the classification task.
    {
        \color{black}
        Following CAE~\cite{Abubakar} and FAE~\cite{wu_fractal_2020}, the discriminative performance is measured by passing the selected features to a downstream classifier~(Extremely Randomized Trees classifier, a variant of Random Forest) as a viable means to benchmark the quality of the selected subset of features.
    }

\begin{figure*}[t]
    \centering
    \includegraphics[width=6.5in]{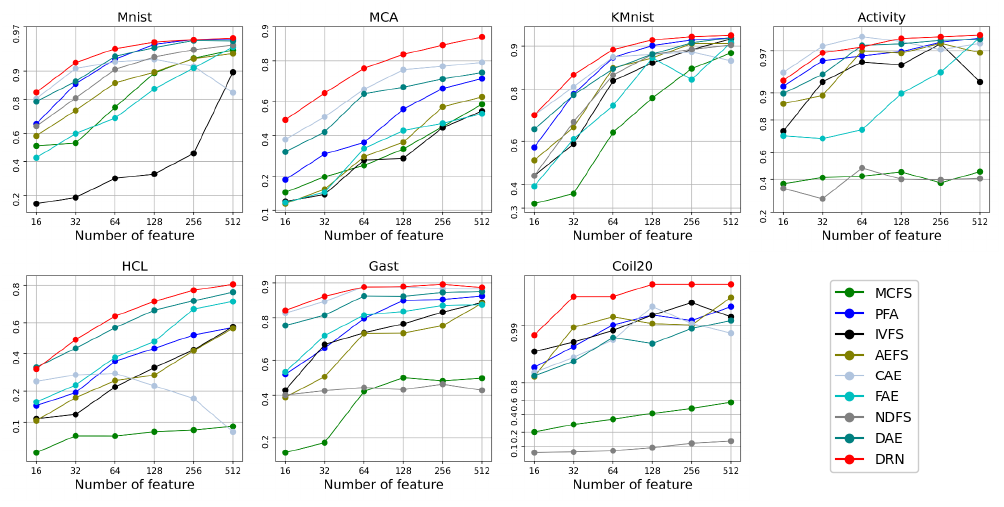}
    % \vspace{-8pt}
    \caption{Classification accuracy with the number of selected features. The horizontal coordinate represents the number of selected features, and the vertical coordinate is the ACC of classification.}
    % \vspace{-8pt}
    \label{fig_graph_clus}
\end{figure*}

For all methods, we select 64 features as benchmarks. The means and standard deviations of the accuracy are shown in Table.~\ref{tab_class_acc_image} and Table.~\ref{tab_class_acc_bio}. 
For a more extensive comparison, we compare the cases of selecting [16, 32, 64, 128, 256, 512] features. The comparison results are shown in Fig.~\ref{fig_graph_clus}.

\textbf{Analysis.}
The conclusions are as follows.
(a) In general, parametric methods are superior to other methods. Among the parametric methods, UDRN has the best results. UDRN has an advantage in all nine datasets. In addition, UDRN outperformed the second-best method by 1\% in six datasets.
(b) UDRN has more advantages in data with more features; for example, in data sets with more than 1000 features~(in Table.~\ref{tab_class_acc_bio}), UDRN has more obvious advantages.

\textbf{Structure-preservation performance.}
The structure-preservation performance tests whether the FS methods preserve the neighborhood relationship of the original data. The structure matching degree~(SMD), a sampling-based structure metric, is chosen as an evaluation metric.
\begin{equation}
    \begin{aligned}
        \text{SMD} = \frac{1}{k |\mathcal{\bm{V}}|}\sum_{i\in \bm{\mathcal{X}}, j\in \mathcal{N}^k(i)} \left|r_{i,j}^\mathcal{\bm{X}}-r_{i,j}^{\mathcal{\bm{X}}^h}\right|
    \end{aligned}
\end{equation}
where the $r_{i,j}^\mathcal{\bm{X}}$ and $r_{i,j}^{\mathcal{\bm{X}}^s}$ are the neighborhood ranking of $\bm{v}_j$ in $\bm{v}_i$ in the input and latent space. The results of the datasets are shown in Table~\ref{tab_SMD}.

\textbf{Analysis.} The conclusions are as follows.
(a) Parametric methods, except UDRN, concentrate on reconstructing all the input features and do not preserve the structure of the selected features well.
(b) Many non-parametric methods design objective functions based on structure retention and achieve suboptimal performance.
(c) UDRN achieves the best score, which is attributed to the fact that the data augmentation of UDRN and the accompanying loss function learn a finer manifold structure.

\begin{table}[tbp]
    % \footnotesize
    \centering
    \caption{ Structure-preserving performance (SMD) comparison, best results are shown in \textbf{blod}.}
    % \vspace{-8pt}
    \begin{tabular}{l|cc|ccc|c}
        \toprule
        {}                    & LS   & IVFS & AEFS & CAE  & FAE  & UDRN           \\ \midrule
        {Coil20}   & 12.0 & 83.2 & 52.5 & 41.5 & 67.8 & {\textbf{86.3}} \\
        {Mnist}    & 17.0 & 86.9 & 59.6 & 42.7 & 46.8 & {\textbf{89.9}} \\
        {KMnist}   & 27.6 & 86.9 & 63.2 & 61.4 & 49.5 & {\textbf{88.3}} \\ 
        {EMnist}   & 31.8 & 69.1 & 71.2 & 75.9 & 65.5 & {\textbf{79.8}} \\ \midrule
        {Activity} & 43.1 & 81.9 & 50.5 & 44.7 & 53.5 & {\textbf{99.9}} \\
        {HCL}      & 11.9 & 22.1 & 11.5 & 10.4 & 13.6 & {\textbf{27.8}} \\
        {Gast}     & 35.2 & 39.0 & 27.0 & 15.8 & 31.2 & {\textbf{48.2}} \\
        {MCA}      & 16.5 & 21.0 & 16.7 & 13.7 & 16.6 & {\textbf{26.9}} \\
        \bottomrule
    \end{tabular}
    % \vspace{-8pt}
    \label{tab_SMD}
\end{table}

\subsection{Comparison with FP methods}
\label{sec_Comparison_with_FP_methods}
This sub-section compares the performance with the baseline FP methods.

For a fair comparison with FP methods, we disable the gate layer by setting $\lambda=0$, these means that all the features can pass the gate layer. Similar to sec.~\ref{sec_Comparison_with_FS_methods}, we evaluate the discriminative performance by the accuracy of the ET tree classifier. The other settings are the same as sec.~\ref{sec_Comparison_with_FS_methods}. The comparison results are shown in Table~\ref{tab_fp}.

\textbf{Analysis.}
The conclusions are as follows.
(a) Except for UDRN, the performance of parametric methods is inferior to that of non-parametric methods. UDRN is not only optimal among all parametric methods but also superior to non-parametric methods. The results are consistent with those in~\cite{Xia2021Revisiting}.
(b) The parametric FP methods are challenging to train because the network parameters need to be optimized rather than the low-dimensional representations. Interestingly, UDRN can solve the training problem of FP networks through data augmentation and novel loss functions.

\begin{table}[tbp]
    \footnotesize
    \centering
    \caption{Discriminative performance comparison with FP methods, best results are shown in \textbf{blod}.}
    % \vspace{-8pt}
    \begin{tabular}{l|cc|ccc|c}
        % \footnotesize
        \toprule
                              & tSNE          & UMAP & GRAE & IVIS & PUMAP & UDRN          \\ \midrule
        \scriptsize{Coil20}   & {78.1}        & 79.6 & 79.9 & 58.9 & 68.9  & \textbf{89.8} \\
        \scriptsize{MNIST}    & \textbf{95.9} & 94.5 & 77.2 & 68.3 & 94.2  & \textbf{95.9} \\
        \scriptsize{KMNIST}   & {64.3}        & 93.8 & 85.4 & 72.8 & 91.5  & \textbf{94.7} \\
        \scriptsize{EMNIST}   & {72.2}        & 74.7 & 72.4 & 36.7 & 77.5  & \textbf{73.5} \\ \midrule
        \scriptsize{Activity} & 92.0          & 91.9 & 88.9 & 82.1 & 90.0  & \textbf{92.6} \\
        \scriptsize{HCL}      & 58.4          & 44.9 & 53.1 & 53.3 & 51.4  & \textbf{91.7} \\
        \scriptsize{Gast}     & 69.5          & 62.7 & 90.1 & 76.4 & 64.5  & \textbf{94.8} \\
        \scriptsize{MCA}      & 51.5          & 41.3 & 83.6 & 66.7 & 46.6  & \textbf{90.9} \\
        \bottomrule
    \end{tabular}
    % \vspace{-8pt}
    \label{tab_fp}
\end{table}

\subsection{Visualization Comparison with FS\&FP Pipeline}

\begin{figure*}[t]
    \centering
    \includegraphics[width=6.7in]{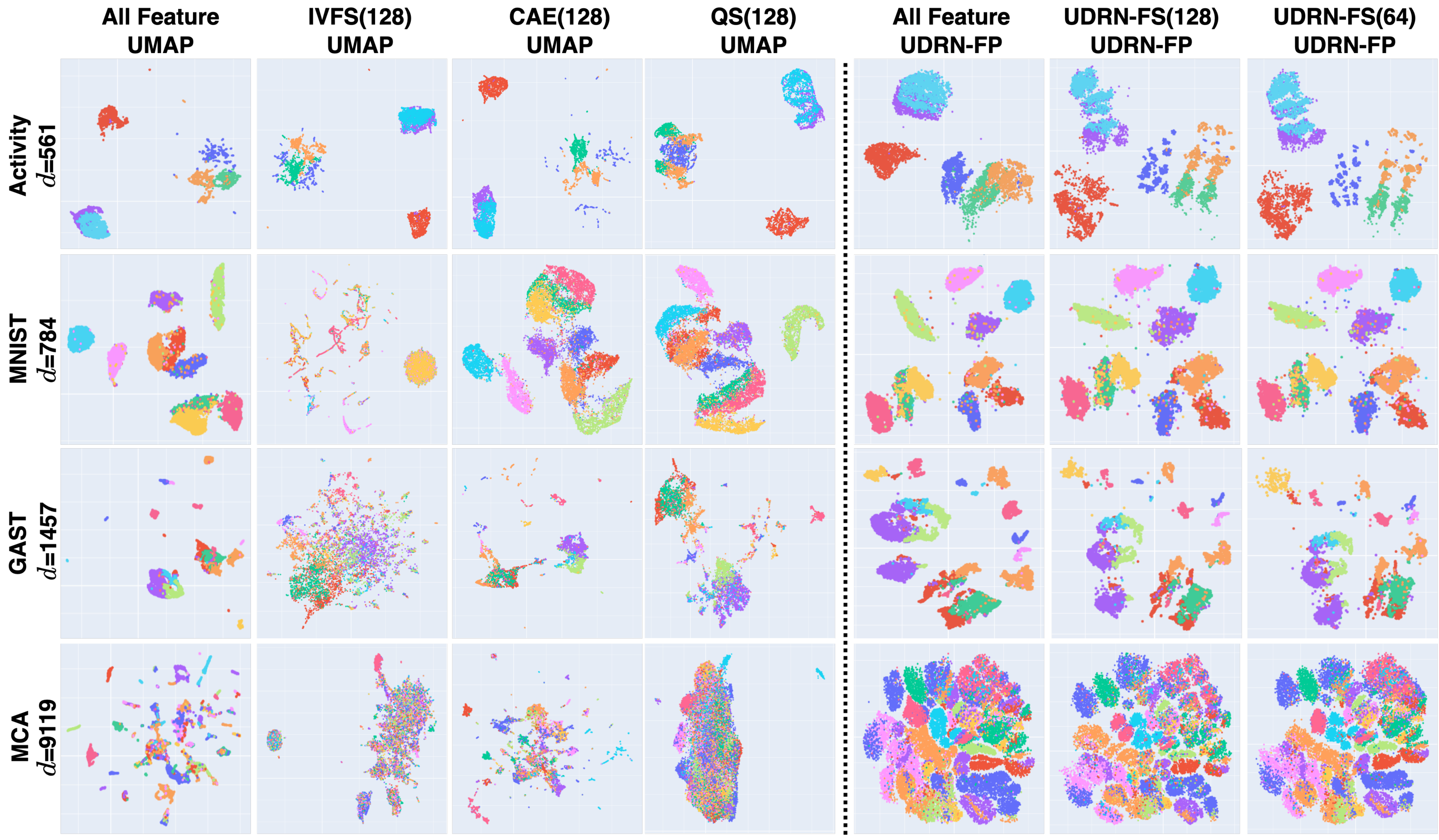}
    % \vspace{-8pt}
    \caption{
        {\color{black} Visualization comparison with FS\&FP pipeline. Each row corresponds to a dataset. In the row names, $D$ represents the number of features in the dataset. Each column corresponds to a combination of FS and FP methods. In the column names, the first occurrence of the feature selection method, the number of selected features is in parentheses, and the second occurrence is the feature mapping method. To align with the baseline nomenclature, our UDRNs are divided into UDRN-FS and UDRN-FP\@. The parentheses after UDRN-FS are used to indicate the number of selected features.}}
    \label{fig_vis}
    % \vspace{-8pt}
\end{figure*}

In application areas such as biology, many articles require a combination of FP and FS methods for data analysis because of the high data dimensionality and excessive noise features~\cite{liang_single_cell_nodate, townes2019feature}.
We have discussed the dilemma of these methods in sec~\ref{sec_relatedwork}. Since the concept of unifying FP\&FS is first proposed by us, these pipeline methods are compared in this article. 
In Fig.~(\ref{fig_vis}), the advantages of UDRN are demonstrated by visualization. The FP method~(UMAP) and FS methods~(IVFS, CAE, and QS), which performed best in sec.\ref{sec_Comparison_with_FS_methods} and sec.\ref{sec_Comparison_with_FP_methods}, are selected as the elements of the pipeline.

\textbf{Analysis.}
The conclusions are as follows.
(a) When using all features, UMAP and UDRN can clearly show the structure, although different in detail.
(b) When selecting fewer features, the compared methods can not guarantee the stability of the data's structure. For example, the embedding of features selected by IVFS differs from the embedding of the original features.
(c) UDRN is very good at producing stable embeddings with only a small number of features due to the uniform loss function and data augmentation. The UDRN performs significantly better than the pipeline method from the visualization point of view.

\subsection{Parameter Analysis}
\begin{table}
    % \centering
    % \footnotesize
    \caption{
        Ablation study. The discriminative performance comparison on the training set with different data augmentation hyperparameters. $0.0$ means without the data augmentation.
    }
    % \vspace{-8pt}
    \begin{tabular}{l|c|cccccc}
        \toprule
        \multicolumn{8}{c}{ Uniform-type Data Aaugmentation, $\tau_{U}(\cdot)$ }                                                            \\ \midrule
        $p_U$        & 0.0  & 0.03          & 0.05          & 0.08          & 0.10          & 0.30          & 0.50          \\ \midrule
        Mnist        & 91.5 & 92.0          & 93.3          & 93.0          & 92.8          & \textbf{93.5} & 93.0          \\
        KMnist       & 84.2 & 88.1          & 87.9          & \textbf{89.5} & 87.6          & 88.6          & 85.7          \\
        EMnist       & 64.7 & 67.5          & 69.8          & 69.5          & \textbf{70.6} & 69.7          & 69.4          \\
        HCL          & 45.2 & 58.8          & 60.1          & 57.1          & 61.0          & 57.8          & \textbf{63.0} \\
        MCA          & 45.5 & \textbf{70.8} & 70.3          & 61.5          & 60.5          & 50.4          & 60.0          \\
        Gast         & 72.0 & 88.4          & 87.6          & \textbf{90.3} & 89.3          & 90.0          & 89.4          \\ \midrule
        AVE          & 68.1 & 78.0          & \textbf{78.2} & 77.2          & 77.3          & 75.3          & 77.6          \\ \midrule
        \multicolumn{8}{c}{ Bernoulli-type Data Aaugmentation, $\tau_{B}(\cdot)$ }                                                          \\ \midrule
        $p_{{B}}$    & 0.0  & 0.03          & 0.05          & 0.08          & 0.10          & 0.30          & 0.50          \\
        \hline Mnist & 91.5 & 93.6          & \textbf{94.6} & 94.2          & 94.1          & 94.0          & 94.0          \\
        KMnist       & 84.2 & 89.7          & 89.7          & 89.7          & \textbf{89.9} & 89.7          & \textbf{89.9} \\
        EMnist       & 64.7 & 71.3          & 71.0          & \textbf{71.4} & 70.9          & 70.4          & 70.2          \\
        {HCL}        & 44.2 & 63.0          & 62.8          & 62.6          & \textbf{63.3} & 62.4          & 59.8          \\
        MCA          & 45.5 & 52.4          & 59.6          & 56.8          & 53.6          & 51.4          & 61.7          \\
        Gast         & 72.0 & 89.4          & \textbf{89.9} & 89.6          & 89.1          & 89.2          & 88.2          \\ \midrule
        AVE          & 68.1 & 77.4          & 77.8          & \textbf{78.4} & 77.9          & 77.0          & 78.0          \\ \midrule
        \multicolumn{8}{c}{ Normal-type Data Aaugmentation, $\tau_{N}(\cdot)$ }                                                             \\ \midrule
        $p_{N}$      & 0.0  & 0.03          & 0.05          & 0.08          & 0.10          & 0.30          & 0.50          \\ \midrule
        Mnist        & 91.5 & 93.8          & 94.1          & 93.8          & \textbf{94.6} & 94.3          & 94.2          \\
        KMnist       & 84.2 & 85.6          & 89.7          & 90.3          & 89.9          & \textbf{90.7} & 90.0          \\
        EMnist       & 64.5 & 65.9          & 69.8          & 69.6          & 70.5          & \textbf{71.4} & \textbf{71.4} \\
        {HCL}        & 47.1 & 56.3          & 56.9          & 57.7          & \textbf{59.8} & 58.9          & 59.4          \\
        {MCA}        & 45.5 & 71.5          & 72.2          & \textbf{75.3} & 74.9          & 73.3          & 73.4          \\
        Gast         & 72.0 & 86.7          & 86.9          & 87.4          & 88.5          & \textbf{90.0} & 89.8          \\\midrule
        AVE          & 68.1 & 77.9          & 78.2          & 78.4          & 79.7          & \textbf{80.1} & 79.4          \\
        \bottomrule
    \end{tabular}
    % \vspace{-8pt}
    \label{tab_parameters_analysis}
\end{table}

% \textbf{Parameter Analysis.}
In this sub-section, the effect of hyperparameters of data augmentation is analyzed and the stability of the hyperparameters is discussed. We follow the sec.~\ref{sec_Comparison_with_FS_methods}'s setup and tested on the Mnist, EMnist, KMnist, HCL, MCA, and Gast datasets. The average ACC is shown in Table.~\ref{tab_parameters_analysis}.

\textbf{Analysis.}
The conclusions are as follows.
(1) FMH augmentation significantly enhances UDRN. Using any data augmentations can dramatically improve the method's performance.
% (2) If the data augmentation is discarded, there is a significant degradation in the performance of UDRN. However, it is not because the proposed loss function cannot bring improvement or is meaningless. The proposed loss function aims to be compatible with data augmentation and becomes a unified whole with data augmentation, so it should be used simultaneously.
(2) The parameters of data augmentation have a relatively small impact on the algorithm, and in short, Normal-type FMH has a relative advantage. In general, the parameters of the FMH augmentation are very stable.

\subsection{Ablation Study}
In this sub-section, the effect of innovations of UDRN is analyzed by ablation experiments. We follow the sec.~\ref{sec_Comparison_with_FS_methods}'s setup and tested on the Mnist, KMnist, EMnist, HCL, and MCA datasets. The discriminative and the structure-preserving performance is shown in Table~\ref{tab_ablation_study} and Table~\ref{tab_ablation_study_structural}. 
\textbf{Ablation 1 (w/o $\tau$)}.
First, the data augmentation $\tau$ is removed. The model is trained directly using the original data $\bm{\mathcal{G}}(\bm{\mathcal{V}}, \bm{\mathcal{E}}, \bm{\mathcal{X}})$.
\textbf{Ablation 2 (w/o $L_\text{tp}$)}.
Second, the proposed loss function $L_\text{tp}$ is removed and replaced with $L_\text{FP}$.
\textbf{Ablation 3 (w/o $\tau$\&$L_\text{tp}$)}.
Finally, both data augmentation $\tau$ and the proposed loss function $L_\text{tp}$ are removed.

\textbf{Analysis.}
The conclusions are as follows.
(1) Both of the innovations presented in this paper lead to performance improvements. And both can work with each other to achieve better performance.
(2) Directly using the reconstructed loss function $L_\text{FP}$ instead of the proposed loss function reduces the performance of the model. It includes both discriminative and structure-preserving performance. Among them, the impact on the structure preservation performance is more significant.
\begin{table}
    \centering
    \caption{Ablation study of UDRN. The discriminative performance comparison under different ablation settings. }
    \begin{tabular}{cllllllll}
        \toprule
                                  & Mnist         & KMnist        & EMnist        & HCL           & MCA           \\ \midrule
        UDRN                      & \textbf{94.3} & \textbf{90.7} & \textbf{71.1} & \textbf{58.9} & \textbf{77.8} \\
        w/o $\tau$                & 91.5          & 84.2          & 64.7          & 44.2          & 45.5          \\
        w/o $L_\text{tp}$         & 87.3          & 85.2          & 63.3          & 35.8          & 56.2          \\
        w/o $\tau$\&$L_\text{tp}$ & 82.5          & 82.3          & 61.8          & 31.6          & 42.1          \\ \bottomrule
    \end{tabular}
    % \vspace{-8pt}
    \label{tab_ablation_study}
\end{table}

\begin{table}
    \centering
    \caption{\color{black} Ablation study of UDRN. The structure-preserving performance comparison under different ablation settings. }
    \begin{tabular}{cllllllll}
        \toprule
        \color{black}                           & \color{black} Mnist         & \color{black} KMnist        & \color{black} EMnist        & \color{black} HCL           & \color{black} MCA           \\ \midrule
        \color{black} UDRN                      & \color{black} \textbf{89.9} & \color{black} \textbf{88.3} & \color{black} \textbf{79.8} & \color{black} \textbf{27.8} & \color{black} \textbf{26.9} \\
        \color{black} w/o $\tau$                & \color{black} {85.3}        & \color{black} {86.2}        & \color{black} {74.3}        & \color{black} {25.5}        & \color{black} {22.7}        \\
        \color{black} w/o $L_\text{tp}$         & \color{black} {35.3}        & \color{black} {39.2}        & \color{black} {25.8}        & \color{black} {15.5}        & \color{black} {17.5}        \\
        \color{black} w/o $\tau$\&$L_\text{tp}$ & \color{black} {36.2}        & \color{black} {34.8}        & \color{black} {26.4}        & \color{black} {17.4}        & \color{black} {12.1}        \\ \bottomrule
    \end{tabular}
    % \vspace{-8pt}
    \label{tab_ablation_study_structural}
\end{table}

\subsection{Supervised UDRN (S-UDRN)}
\begin{figure}[t]
    \centering
    \includegraphics[width=3.0in]{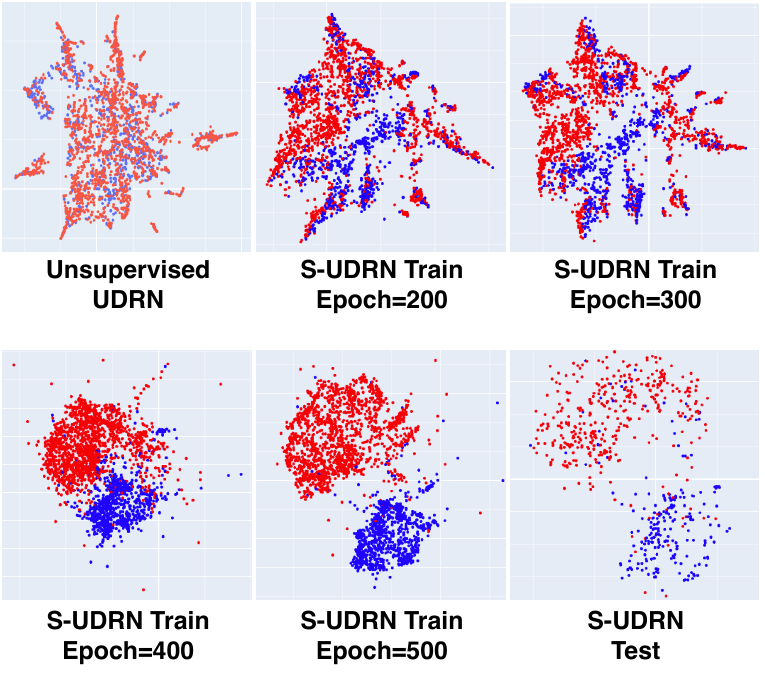}
    % \vspace{-8pt}
    \caption{Visualization of unsupervised UDRN~(UDRN) and supervised UDRN~(S-UDRN). The proposed UDRN method is also compatible with the supervised case and performs FS\&FP based on data labels or other prior knowledge.}
    % \vspace{-8pt}
    \label{fig_sup}
\end{figure}

UDRN is well compatible with both supervised and unsupervised cases and only needs to consider whether supervised information is available when constructing the graph structure (in Eq.~(\ref{eq_unsup}) and Eq.~(\ref{eq_sup})).
We chose a gut flora dataset to test UDRN under supervised scenarios. Since the gut flora dataset has few features associated with the labels, UDRN is needed to select highly relevant features from all the labels and complete the FP (in Fig.~\ref{fig_sup}).

\textbf{Analysis.}
(a) The dataset contains many useless features such that the unsupervised scheme is unable to distinguish classes.
(b) Our UDRN approach can learn to distinguish labeled features based on a modified graph structure (based on labels), while at the same time, the intra-class manifold structure embedding does not receive influence.

\section{Conclusion}

We developed an integrated method for feature selection (FS) and feature projection (FP) with the help of neural networks named Unified Dimensional Reduction Neural-networks~(UDRN). UDRN handles FS and FP tasks end-to-end with a unified loss function.
UDRN is compatible with both supervised and unsupervised settings. We demonstrate the effectiveness and sophistication of UDRN by working with state-of-the-art FS, FP, and FS\&FP pipeline methods.

% \appendix
% \section{My Appendix}
% Appendix sections are coded under \verb+\appendix+.

% \verb+\printcredits+ command is used after appendix sections to list 
% author credit taxonomy contribution roles tagged using \verb+\credit+ 
% in frontmatter.
\clearpage
\printcredits

\bibliographystyle{model1-num-names}
% \bibliographystyle{cas-model2-names}

% \bibliography{cas-refs}
\bibliography{zzl_lib,zzl_mldl,zotero}

\end{document}